\documentclass[preprint,12pt]{elsarticle}
\usepackage{amssymb}
\usepackage[utf8]{inputenc} %
\usepackage[T1]{fontenc}    %
\usepackage{hyperref}       %
\usepackage{url}            %
\usepackage{booktabs}       %
\usepackage{amsfonts}       %
\usepackage{nicefrac}       %
\usepackage{microtype}      %
\usepackage{paralist}
\usepackage{microtype}
\usepackage{graphicx}
\usepackage{subcaption}
\usepackage{booktabs} %
\usepackage{amsfonts,amsmath,amssymb,amsthm}
\usepackage{bm,nicefrac}

\usepackage{floatrow}
\newfloatcommand{capbtabbox}{table}[][\FBwidth]

\usepackage{algpseudocode}
\usepackage{algorithm}
\usepackage{lineno,hyperref}
\usepackage{lipsum}  

\usepackage[capitalize]{cleveref}
\usepackage[shortcuts]{extdash}
\usepackage[all]{foreign}
\usepackage{siunitx}

\usepackage{tikz}
\usetikzlibrary{decorations.pathreplacing}

\algnewcommand\algorithmicforeach{\textbf{for each}}
\algdef{S}[FOR]{ForEach}[1]{\algorithmicforeach\ #1\ \algorithmicdo}
\usepackage{xspace}
\newcommand{\edl}{\mbox{\textsf{EDL}}\xspace}

\usepackage{array}
\newcolumntype{?}{!{\vrule width 1.2pt}}

\usepackage[]{graphicx,amssymb}    %
\usepackage{amsmath,bm,amsthm}
\usepackage{epsfig}
\usepackage{epstopdf}
\usepackage{color}
\usepackage{caption}
\newcommand{\Dir}[1]{\textnormal{Dirichlet} (#1)}
\newcommand{\KL}[2]{\textnormal{KL}[#1\|#2]}

\newtheorem{proposition}{Proposition}

\journal{}

\newcommand{\desideratum}[1]{\mbox{\textbf{D#1}}}
\newcommand{\desdirichlet}{\desideratum{1}}
\newcommand{\destransfer}{\desideratum{2}}
\newcommand{\descompositionality}{\desideratum{3}}

\usepackage{float}
\newfloat{myalgorithm}{t}{lop}
\floatname{myalgorithm}{Algorithm}

\begin{document}

\begin{frontmatter}

\title{Risk-aware Classification via Uncertainty Quantification}

\author[inst1]{Murat Sensoy}
\ead{msensoy@amazon.co.uk}
\author[inst2]{Lance M. Kaplan}
\ead{lance.m.kaplan.civ@army.mil}
\author[inst3]{Simon Julier} 
\ead{s.julier@ucl.ac.uk}
\author[inst4]{Maryam Saleki}
\ead{msaleki@fordham.edu}
\author[inst5,inst6]{Federico Cerutti\corref{cor1}}
\ead{federico.cerutti@unibs.it}

\cortext[cor1]{Corresponding author}

\affiliation[inst1]{organization={Amazon Alexa AI},
            city={London},
            postcode={EC2A 2FA},
            country={UK}}
\affiliation[inst2]{organization={DEVCOM Army Research Laboratory},%
            city={Adelphi}, 
            state={MD},
            country={US}}
\affiliation[inst3]{organization={University College London}, 
            postcode={WC1E 6BT},
            city={London},
            country={UK}}
\affiliation[inst4]{organization={Fordham University},
            city={New York},
            country={USA}} 

\affiliation[inst5]{organization={University of Brescia}, 
            city={Brescia},
            country={Italy}}
\affiliation[inst6]{organization={Cardiff University}, 
            city={Cardiff},
            country={UK}}

\begin{abstract}
Autonomous and semi-autonomous systems are using deep learning models to improve decision-making. However, deep classifiers can be overly confident in their incorrect predictions, a major issue especially in safety-critical domains.
The present study introduces three foundational desiderata for developing real-world risk-aware classification systems. Expanding upon the previously proposed Evidential Deep Learning (\edl), we demonstrate the unity between these principles and \edl's operational attributes. 
We then augment \edl empowering autonomous agents to exercise discretion during structured decision-making when uncertainty and risks are inherent.
We rigorously examine empirical scenarios to substantiate these theoretical innovations. In contrast to existing risk-aware classifiers, our proposed methodologies consistently exhibit superior performance, underscoring their transformative potential in risk-conscious classification strategies.
\end{abstract}


\end{frontmatter}

\section{Introduction}
\label{sec:intro}
In recent years, we have witnessed unprecedented advances and breakthroughs in machine intelligence, where deep learning has played an important role.
While it allows neural networks to achieve near (or even super)human performance in many tasks, such as object classification, the resulting models are usually overconfident when they make mistakes \cite{guo2017calibration}. 
These models may have high test set accuracy; however, they cannot be used easily in critical settings where the cost of one error may significantly surpass the benefit of their success in many others.
In 2016, an autonomous car collided with a truck. The car's driver lost their life due to a vision system error, which misidentified the truck as a bright sky, leading to the accident \cite{NHTSA}.

With the increasing prevalence of deep learning in critical decision-making scenarios, the challenges linked to misclassification are poised to escalate, underscoring the necessity for the development of \textit{risk-aware classification systems}.
In our paper, we put forth three desiderata for risk-aware classification machinery, \cf \Cref{sec:motivation_full_section}: uncertainty quantification with multinomial's conjugate distribution; effective knowledge transfer from pre-trained models; and compositionality capabilities.
This is fundamentally important when assessing pignistic probabilities \cite{smets_DecisionmakingTBM_05} in the real world, \ie the likelihood that a rational agent will opt for a specific choice when faced with a decision-making scenario. Pignistic probabilities encompass the uncertainty faced by the decision maker when confronted with many options along with the associated risk associated with selecting each option. Moreover, pre-trained models are standard in modern machine learning applications, as they reduce development time and resource requirements, therefore we must support the smooth transfer of knowledge from these models to maintain functionality and adaptability. Finally, compositionality enables handling disjoint categories by integrating separate classifiers, another common, resource-efficient approach used in modern machine learning applications.

In \Cref{sec:edl_learning}, we revisit the \edl proposal \--- introduced initially at the NeurIPS conference in 2018 \cite{edl} by some of the authors of this paper \--- to show that it fulfills the three desiderata. Not only that, it does it simply and elegantly, without \--- for instance \--- requiring external datasets. 
In particular, we demonstrate that the \edl loss function can be used to fine-tune pre-trained classifiers to improve their uncertainty quantification and turn them into \edl classifiers.
We also propose a principled approach for fusing evidential classifiers trained for different categories and datasets.

In Section \ref{sec:decision}, we extend \edl by incorporating a principled assessment of the misclassification risk, significantly expanding the preliminary results presented at  \cite{DBLP:conf/wacv/SensoySJAR21}, and show how risk-aware \edl classifiers can be trained in environments with bandit feedback.
More specifically, risk-awareness is enforced through Dirichlet priors, which take precedence only when the model is uncertain.
Hence, the model chooses the most likely category for classification when it is confident but refrains from making risky classification decisions when it is not. 

Our extensive experimental analysis (Section \ref{sec:evaluation}) demonstrates that our approaches significantly improve the original \edl proposal regarding cost minimization and outperform the state-of-the-art cost-sensitive training approach for deep learning models.
We also discuss the most recent work on uncertainty quantification and risk-aware classification in \Cref{sec:related}.

\section{Motivation and Desiderata for Real-World Risk-Aware Classification}
\label{sec:motivation_full_section}
As we show in \Cref{sec:softmax}, due to the nature of the softmax activation function and the inherent closed-world assumption in classification problems, out-of-distribution samples can easily be confidently misclassified.
The standard practice for deep neural networks is to utilize the softmax function to transform the continuous output layer activations into class probabilities. 

By interpreting, instead, a classification as a realization of a Dirichlet distribution, \cf \Cref{sec:dir}, we can estimate the associated epistemic uncertainty. Epistemic uncertainty emerges due to the limitations in the agent's model knowledge and can potentially be mitigated by gathering additional data samples \cite{HORA1996217,hullermeier_AleatoricEpistemicUncertainty_21,cerutti_EvidentialReasoningLearning_22a}.

The Dirichlet distribution \--- being conjugate to the multinomial distribution \--- yields a posterior that is also a Dirichlet distribution when employed as a prior in weighting evidence for the various classes. Therefore, its variance provides an estimate of the associated epistemic uncertainty. As more evidence is collected, the lower the uncertainty becomes. As a result, when the sample sits in a close neighbourhood of analogous samples which have been observed multiple times in the training phase, the variance of the posterior distribution will be very small.

The Dirichlet distribution provides a mathematical framework for representing pignistic probabilities in the presence of risky classifications, \ie costly misclassification, such as false negative tests for infectious disease. In \Cref{sec:desiderata}, we discuss how pignistic probabilities \cite{smets_DecisionmakingTBM_05} depict the likelihood that a rational agent will opt for a specific choice when faced with a decision-making scenario. We also distil the desiderata for devising an effective real-world risk-aware classification machinery, which we anticipate here:
\begin{description}
\item[\desdirichlet]: the output of a classification system should be interpreted as a Dirichlet distribution with some estimation of its epistemic uncertainty;
\item[\destransfer]: the classification system should ensure the possibility to transfer knowledge from existing, pre-trained models smoothly;
\item[\descompositionality]: the classification system should show some forms of compositionality capabilities, \ie it should be possible to fuse the predictions of two classification systems on different categories.
\end{description}

These three desiderata are necessary for the development of real-world risk-aware classifiers, which need to address critical challenges such as over-confident predictions, efficient model utilisation, and flexible handling of diverse categories. We, however, do not claim that these requirements are sufficient. While they address several important aspects, they might not cover all potential challenges and needs in the evolving field of risk-aware classification. As the field progresses, new challenges and requirements will likely emerge. Potential additional requirements might include adaptability to new data distributions, scalability to handle large-scale data, providing interpretable uncertainty estimates, robustness to adversarial attacks, and optimising for cost and resource efficiency. These additional factors could further enhance the effectiveness and resilience of risk-aware classifiers in real-world applications. We further comment on this aspect in \Cref{sec:conclusions}.

\subsection{The Problem with the Over-Confident Classifier}
\label{sec:softmax}
Given an observed tuple $({\bm x}, {y})$, 
the likelihood function for a $K$-class classification problem is
\begin{align*}
   Pr(y|{\bm x},\theta) = \mathrm{Mult}(y|\sigma_1(f_{\theta}({\bm x})), \cdots, \sigma_K(f_{\theta}({\bm x}))),
\end{align*}
such that $\mathrm{Mult}(\cdots)$ is a multinomial mass function, $f_{\theta}(\bm{x})$ can be approximated with a neural network with parameters $\theta$ and $\sigma_j(\bm{u})=\frac{e^{u_j}}{\sum_{i=1}^K e^{u_K}}$ is the projection function extracting the $j^{th}$ element, in this case, of the softmax function.%

The maximization of the multinomial likelihood with respect to the neural network parameters $\theta$ is accomplished with a preference for the computational convenience offered by the equivalent problem of minimizing the negative log-likelihood
\begin{align*}
 -\log p(y|{\bm x},\theta) = -\log \sigma_y(f_{\theta}({\bm x}))
\end{align*}
also named {\it the cross-entropy loss}.
\begin{figure}[t!]
$\begin{array}{c}
    \includegraphics[width=6.3cm]{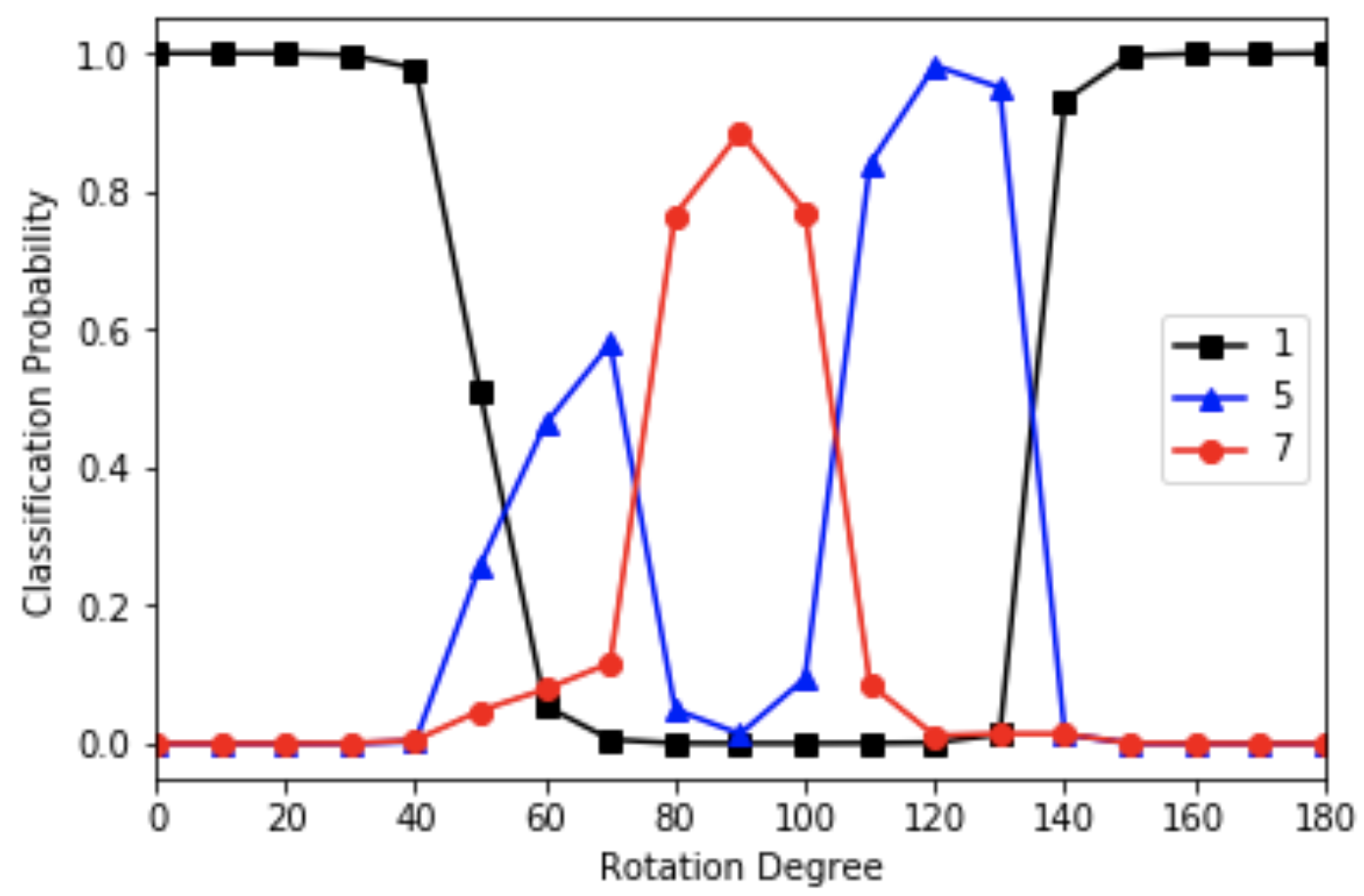} \\%& \includegraphics[width=6.3cm]{figs/new/edl_rotate.png}\\
    \multicolumn{1}{c} {~~~~~~\fbox{\includegraphics[width=5.5cm]{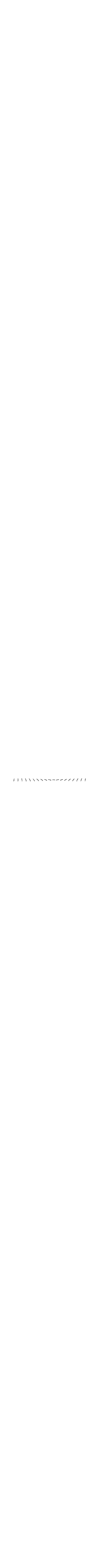}}} \\%& \multicolumn{1}{c} {~~~~~~\fbox{\includegraphics[width=5.5cm]{figs/nips18/rotating.pdf}}}
\end{array}$
\caption{\label{fig:softmax} Classifying a rotated digit $1$ (positioned at the bottom) across various angles spanning from 0 to 180 degrees involves the calculation of classification probabilities through the \textit{softmax} function.}
\end{figure}
Significantly, the cross-entropy loss's probabilistic interpretation is solely based on Maximum Likelihood Estimation (MLE). MLE \--- being a frequentist approach \--- lacks the capability to deduce the variance of the predictive distribution. 
The outcome of this behavior is an unreliable estimation of uncertainty. 

Inspired by~\cite{gal2016icml} and \cite{Louizos17}, Figure~\ref{fig:softmax} illustrates the failure of LeNet~\cite{LeCun1999} in classifying a depiction \--- from the MNIST dataset \--- of the digit $1$ as it undergoes counterclockwise rotation. LeNet, along with numerous other architectures, relies on the softmax function to estimate classification probabilities. However, as the image undergoes rotation, it struggles to provide accurate classifications.
For slight degrees of rotation, the result is a correct and confident classification. However, as the image undergoes rotation between $70$ and $100$ degrees, it switches to being confidently classified as $7$. Further rotation between $110$ and $130$ degrees leads to categorizing the image as $5$ confidently.

The desired behaviour, instead, would be for the classifier to acknowledge that images rotated between $60$ and $130$ degrees are out of the distribution considered in training. As mentioned, the Dirichlet distribution provides formal machinery for expressing such epistemic uncertainty.

\subsection{A Primer in Dirichlet Distribution}

\label{sec:dir}
The Dirichlet distribution \--- being a multivariate generalization of the beta distribution \--- is a probability distribution that is commonly used to model the distribution of proportions or probabilities over $K$ categories $\bm{\pi} =[\pi_1,\ldots, \pi_K]$:
\begin{equation} \nonumber
\Dir{\bm{\pi}|\bm{\alpha}} = \left \{ \begin{array}{ll}
\frac{1}{B(\bm{\alpha})} \prod_{i=1}^K \pi_i^{\alpha_i-1} & \mbox{for $\bm{\pi} \in \mathcal{S}_K $},\\
0 & \mbox{otherwise}. \end{array} \right.
\end{equation}
where 
$\bm{\alpha}=[\alpha_1, \ldots, \alpha_K]$ are its parameters, 
$\mathcal{S}_K$ is the $K$-dimensional unit simplex, and $B(\bm{\alpha})$ is the $K$-dimensional multinomial beta function~\cite{kotz.00}.
The mean ($\bar{\pi}_k$) and variance of the probability of category $k$ are
\begin{equation}\label{eq:dir-moments}
  \bar{\pi}_k = \mathbb{E}[\pi_k] = \frac{\alpha_k}{\alpha_0},
  ~~
  \mathbb{V}[\pi_k] = \dfrac{\alpha_{k} (\alpha_0-\alpha_{k})}{\alpha_0^2 (\alpha_0 + 1) }\text{, where }\alpha_0=\sum_{i=1}^K \alpha_i.
\end{equation}

\begin{figure*}
\centering
    \begin{subfigure}[b]{0.3\textwidth}
        \includegraphics[width=\textwidth]{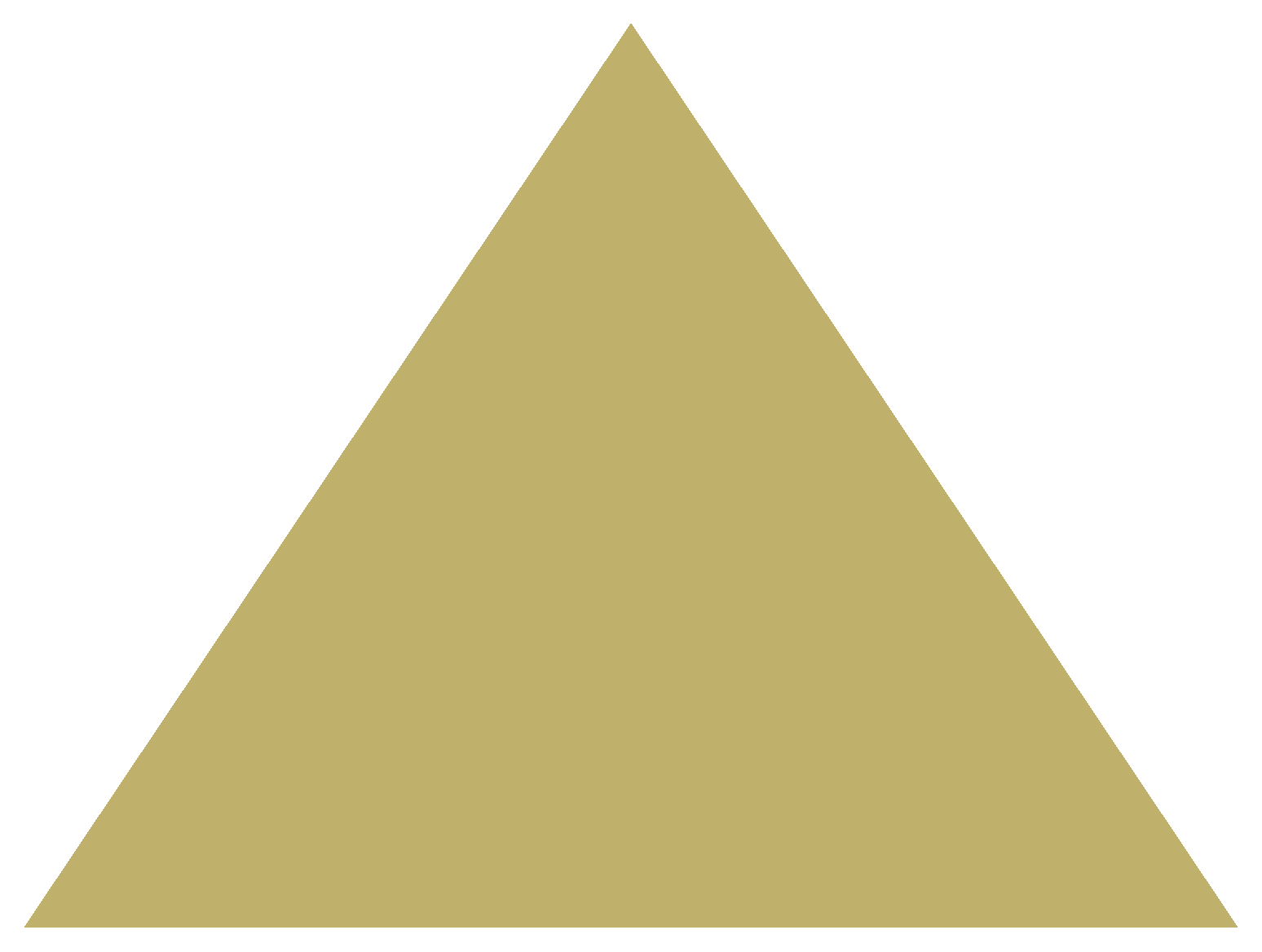}
        \caption{$\bm{\alpha}=[1,1,1]$}
        \label{fig:dirichlet_1_1_1}
    \end{subfigure}
    \hfill
    \begin{subfigure}[b]{0.3\textwidth}
        \includegraphics[width=\textwidth]{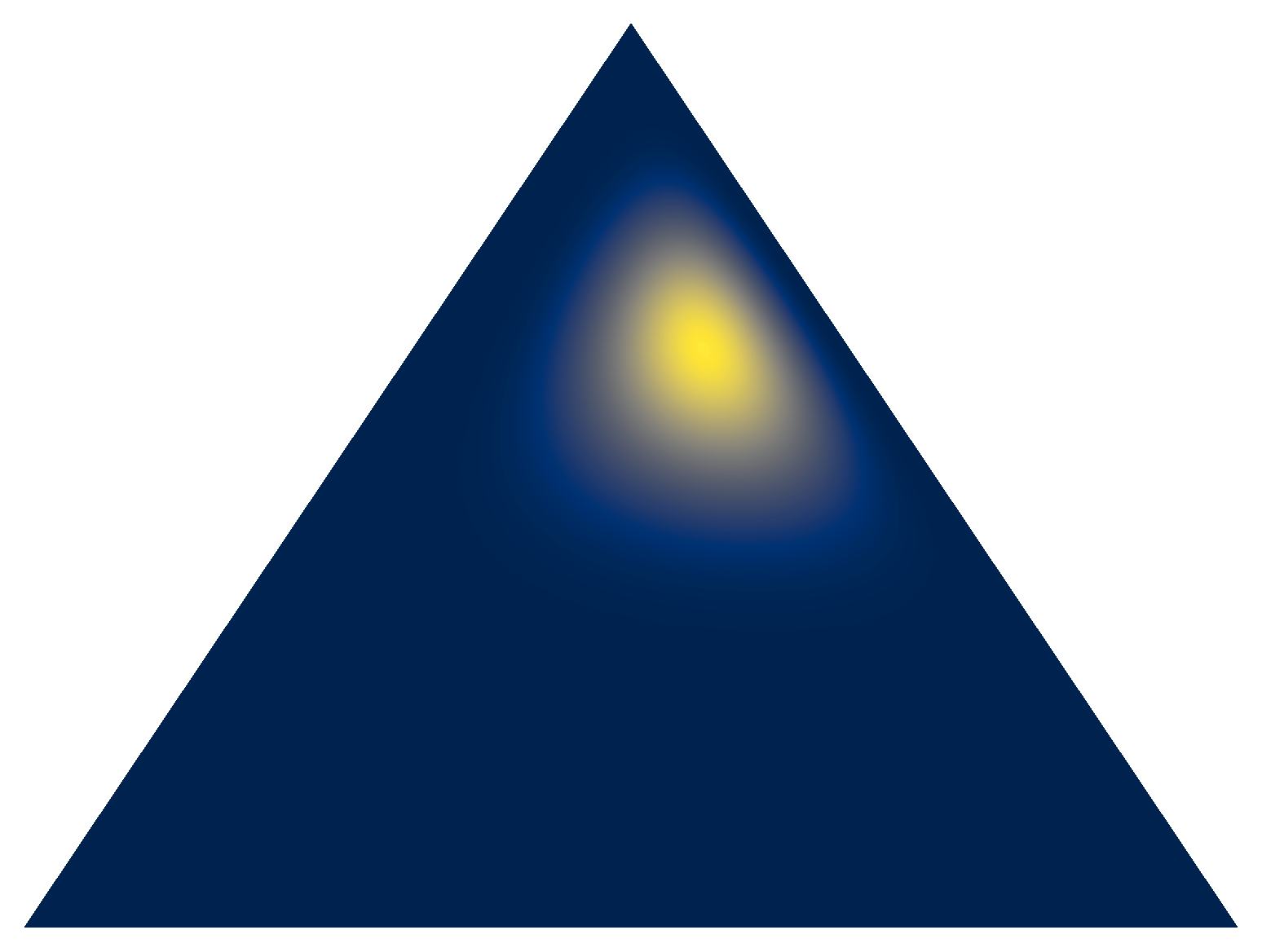}
        \caption{$\bm{\alpha}=[4,7,17]$}
        \label{fig:dirichlet_4_7_17}
    \end{subfigure}
    \hfill
    \begin{subfigure}[b]{0.3\textwidth}
        \includegraphics[width=\textwidth]{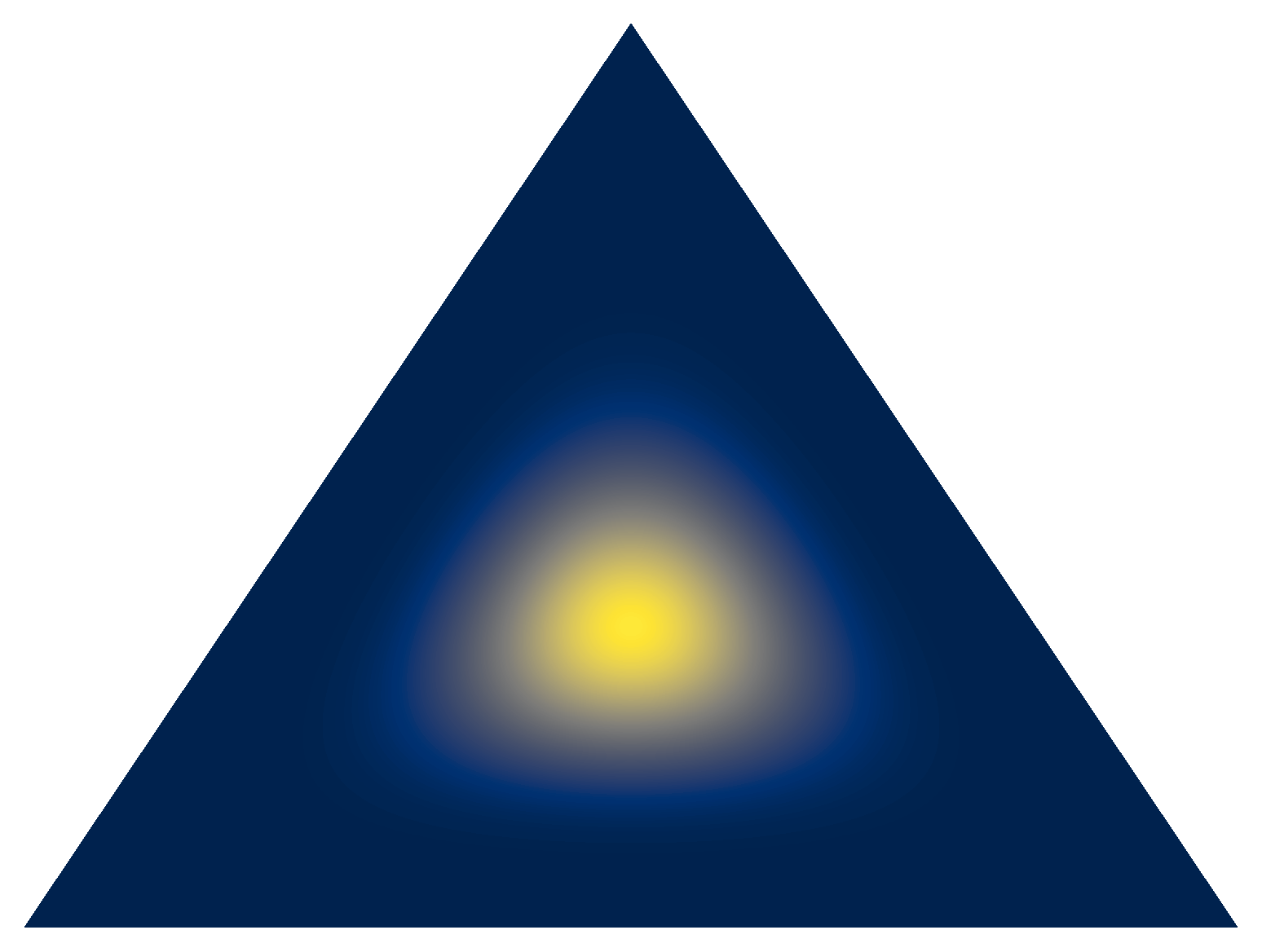}
        \caption{$\bm{\alpha}=[8,8,8]$}
        \label{fig:dirichlet_8_8_8}
    \end{subfigure}
\caption{\label{fig:dirichlets}
Probability density function of different $\Dir{\bm{\pi}|\bm{\alpha}}$ distributions over three possible outcomes.}
\end{figure*}

\Cref{fig:dirichlets} visually depicts various Dirichlet distributions with varying parameter values over three possible outcomes. When all parameters are set to one (\ie $\bm{\alpha}=[1,1,1]$, \cf \Cref{fig:dirichlet_1_1_1}), the Dirichlet distribution converts to a uniform distribution.

The Dirichlet parameters can be conceptually regarded as real-valued pseudocounts~\cite{murphy2012machine}. In this interpretation, higher pseudocount values signify stronger evidence in favour of a particular category. This interpretation becomes more rigorous when seeing it as the conjugate prior of a categorical distribution. Because of the conjugacy property, the posterior distribution \--- when observing pseudocounts of evidence $\bm{c} = [c_1, \ldots, c_K] \sim \text{Multinomial}(n, \bm{\pi})$ \--- becomes $\Dir{\bm{\pi}| \bm{\beta} + \bm{c}}$, with $\Dir{\bm{\pi}; \bm{\beta}}$ our prior belief. For instance, if we consider $\bm{c} = [3, 6, 16]$ as the evidence to be incorporated into the initial pseudocounts $\bm{\beta} = [1,1,1]$, the posterior distribution is $\Dir{\bm{\pi}; [4, 7, 17]}$, hence, the third category is now the most probable one, \cf \Cref{fig:dirichlet_4_7_17}.
However, it is important to note that more evidence does not necessarily lead to one category becoming the most probable. For example, consider the case that $\bm{c} = [7, 7, 7]$. The posterior's parameters are $\bm{\beta} = [8, 8, 8]$, suggesting that all categories remain equally probable. However, as shown in \Cref{fig:dirichlet_8_8_8}, the variance of the posterior is considerably smaller.

The Dirichlet distribution provides us with a mathematical framework for representing pignistic probabilities in the presence of risky classifications, \ie costly misclassification, such as false negative tests for infectious disease.

\subsection{Real-World Risk-Aware Classification Desiderata}
\label{sec:desiderata}

Let us now consider an autonomous agent that must classify samples \--- \eg medical CAT scans in search of possible anomalies \--- into $K$ distinct categories.
The agent's classification decision for a given sample may involve some cost if it is wrong. This cost is referred to as \textit{risk} in the rest of the paper.

The risk of misclassification may be task-specific and agent-specific. It can be encoded compactly into an asymmetric non-negative square risk matrix, which is written as $\bm{R} \in {[0, \infty)}^{K \times K}$.  %
The $k^{\text{th}}$ row of this matrix is the vector $\bm{R}_k$, whose elements $R_{ki}$ indicate the risk of classifying a sample from category $k$ as category $i$. The risk of correct classification is zero, \ie $R_{kk}=0$ for any $k$ and $R_{ki} \gg R_{kk}$ for all $i \neq k$.
In order to minimize costs associated with classification decisions, an agent must learn to classify samples using its risk matrix properly. This means that the agent may choose not to classify a sample in the category with the highest probability if the misclassification cost is too high for that category.

Pignistic probabilities \cite{smets_DecisionmakingTBM_05} capture the likelihood of a specific choice when facing a decision-making scenario. These pignistic probabilities are denoted as $\bm{p}=[p_1,\ldots, p_K]$ and hold an intrinsic mathematical equivalence to Shapley's value \cite{dubois2008definition}. They encompass the uncertainty faced by the decision maker when confronted with many options, \ie the uncertainty pertaining to $\bm{\pi}$, along with the associated risk associated with selecting each option.
Therefore, pignistic probabilities ($\bm{p}$) are employed \--- instead of, for instance, categorical probabilities $\bm{\pi}$ \--- when calculating the expected risk of a classification. This substitution is carried out to factor in the risk of misclassification appropriately.

It appears vital that \textbf{the output of a classification system should be interpreted as a Dirichlet distribution} with some estimation of its epistemic uncertainty (\desdirichlet) because the epistemic uncertainty is a fundamental aspect to consider in determining the final choice among the $K$ options. 
Moreover, to ensure the possibility of realistic applications, we need to identify the \textbf{simplest} method that satisfies \desdirichlet. Simplicity here concerns mainly with the ability to enable risk awareness on existing classification systems. 

\textbf{Pre-trained models} are standard in modern machine learning applications, as they reduce development time and resource requirements. Retraining such models from scratch is often infeasible due to the computational costs they incur. Therefore, a classification system must support the \textbf{smooth transfer of knowledge from these models} (\destransfer) to maintain functionality and adaptability. A practical approach to reusing legacy classifiers is to replace the penultimate layer with a new head that outputs a Dirichlet distribution for classification. This method is resource-efficient, as it capitalises on the pre-existing feature extraction capabilities of the model while updating only the classification component.

Finally, we wish for risk-aware classifiers to support effective decision-making in the real world, which is continuously changing. A necessary characteristic is, therefore, the possibility \--- for the classification system \--- \textbf{to manifest some ability of compositionality} (\descompositionality), \ie it is desirable to be able to fuse the predictions of two classification systems on different sets of categories. 
Compositionality enables handling disjoint categories by integrating separate classifiers, a common, resource-efficient approach used in modern machine learning applications. This can improve accuracy and robustness, by leveraging their complementary strengths and balancing diverse error patterns, as each model's unique capabilities and mistakes can offset those of the other.
Compositionality also supports incremental updates, allowing new classifiers to be added without disrupting the existing system, and is more resource-efficient by using specialised, smaller models. 

In the next section, we expand upon Evidential Deep Learning (\edl) \cite{edl} \--- initially proposed by some of the authors of this paper \--- and show how to satisfy the three desiderata listed above.

\section{Evidential Deep Learning (\edl) Revisited}
\label{sec:edl_learning}
\begin{figure}
    \centering
    \includegraphics[height=3.5cm]{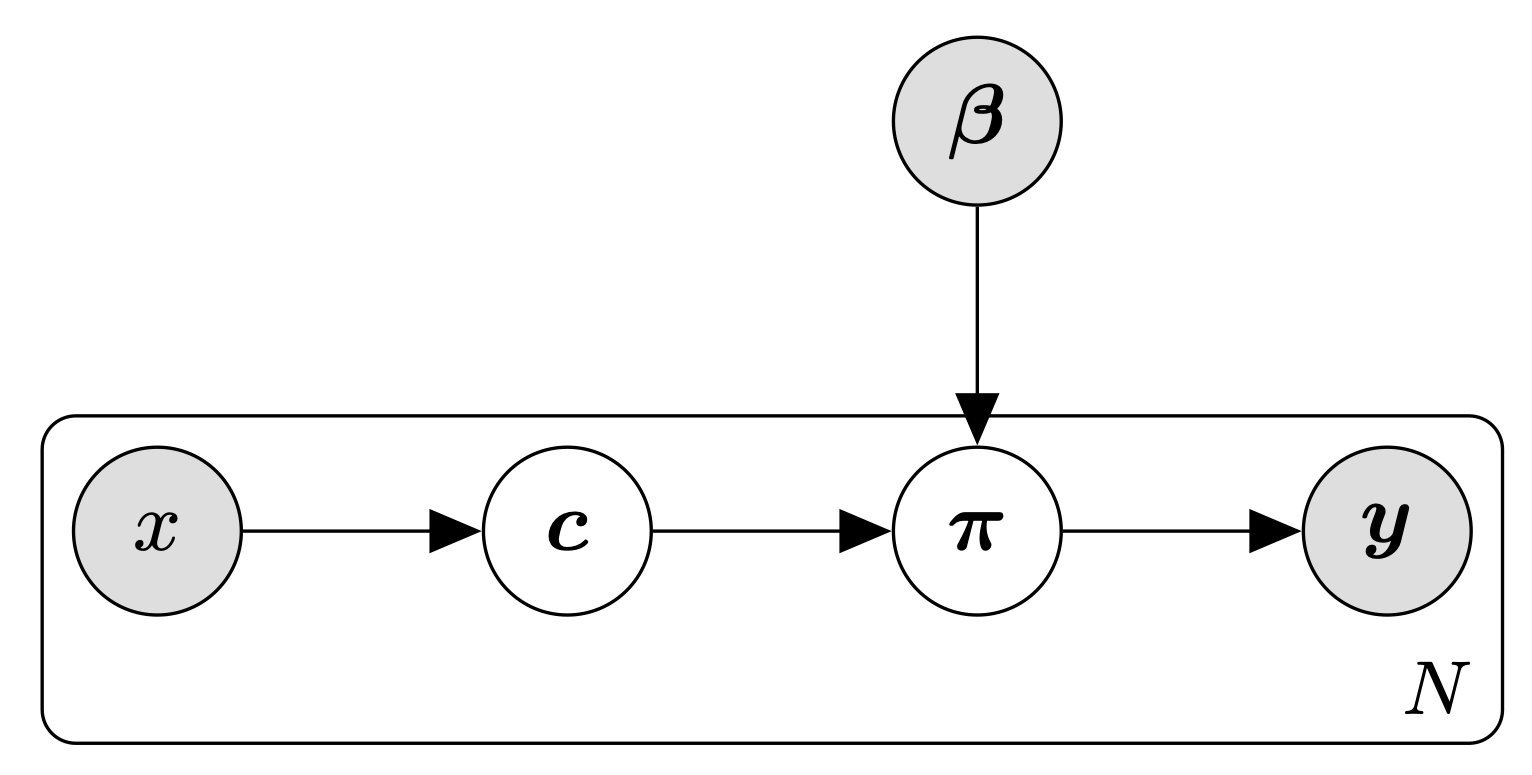}
    \caption{\label{fig:disc_model}Visual depiction of the plate notation for discriminative evidential classifiers.}
\end{figure}

\Cref{fig:disc_model} illustrates a generic discriminative classifier \cite{murphy2012machine}, which directly estimates a distribution $p(y|\bm{x})$ over $K$ possible categories.
It operates under the assumption that each input sample $\bm{x}$ carries with it a pseudo-count of evidence denoted as $\bm{c} = [c_1, \dots, c_K]$, where $c_i > 0$ for all $i$. 
These $c_i$ values effectively serve as pseudocounts, quantifying the strength of evidence in favor of $i^\text{th}$ category. This evidential information is then integrated with a prior distribution to parameterize a distribution over a latent variable unique to each sample, denoted as $\bm{\pi} \sim \Dir{\bm{\pi}|\bm{\alpha}}$, where $\bm{\alpha} = \bm{\beta} + \bm{c}$. The label $y$ is thus sampled from this $\bm{\pi}$ distribution.

\edl estimates a probability distribution \--- not a point estimation \---  over $\bm{\pi}$.
Despite $\bm{\pi}$ being a latent variable, it is possible to perform marginalization over it, \ie $p(y=k|\bm{c}, \bm{\beta}) = p(y=k|\bm{\alpha}) = \alpha_k/\alpha_0$, \cf \Cref{eq:dir-moments}.

For any given input $\bm{x}$, the prior $\bm{\beta}$ \--- in the following fixed to $[1, \ldots, 1]$, \cf \eg \Cref{fig:dirichlet_1_1_1},  to ensure a uniform distribution as a prior \--- is combined with the evidence $\bm{c}$ to estimate the distribution of $\bm{\pi}$.
The entropy of the predictive categorical distribution is then:
\begin{equation}
    \mathbb{H}[y|\bm{\alpha}] = - \sum_{i=1}^K \frac{\alpha_i}{\alpha_0} \log \frac{\alpha_i}{\alpha_0}
\end{equation}
Here, $\alpha_0=\sum_{i=1}^K \alpha_i$, as outlined in \Cref{eq:dir-moments}.
When the evidence is limited, the prior will exert greater influence, resulting in heightened uncertainty in the value of $y$. Conversely, if the evidence is substantial and concentrated in a particular category, the uncertainty will diminish.

\edl \cite{Sensoy2018} is a very efficient proposal as it does not rely on external sources of knowledge \--- \eg additional datasets \--- for ascertaining Dirichlet distributions instead of point probabilities. 
Nowadays, there are many proposals that explore this direction, as evidenced by recent surveys \cite{hullermeier_AleatoricEpistemicUncertainty_21,cerutti_EvidentialReasoningLearning_22a,abdar_reviewuncertaintyquantification_21}. 
In \Cref{sec:related}, we comment on some of the other approaches that could \--- in principle \--- still be used for risk-aware classification. However, at the time of writing no publications provide the evidence that they can satisfy \--- efficiently \--- all the desiderata we discussed in \Cref{sec:motivation_full_section}. We, instead, demonstrate it for \edl in the following sections.

\edl distinguishes itself from standard neural networks by substituting the conventional \textit{softmax} layer with a non-negative activation layer, which is regarded as the evidence vector for predicting the Dirichlet distribution. In essence, when provided with the neural network output $f_{\theta}(\bm{x}_i)$ from the logits layer for input $\bm{x}_i$, the evidence vector is computed as $c_{\theta}({\bm x}_i) = \zeta({f_{\theta}(\bm{x}_i)})$, where $\zeta(\cdot)$ represents an element-wise non-negative function such as \textit{ReLU}, \textit{Exponent}, or \textit{Softplus}.
Such results can be the base for the parameters  $\bm{\alpha}_i = c_{\theta}({\bm x}_i) + 1$ of a Dirichlet distribution: their mean, $\bm{\alpha}_i/\sum_{k=1}^{K} \alpha_{ik}$, can be employed as an estimation of the class probabilities.

Hence, to transform an existing neural network classifier into an \edl, only two changes are necessary:
\begin{enumerate}
    \item replace the output to derive a vector of non-negative real values. These different options of activation function as hyper-parameters and the selection should be based on the network architecture, problem, and dataset similar to other hyper-parameters such as learning rate. In the following, we followed the standard procedure for determining such hyper-parameters through empirical analysis over a validation set;
    \item modify the loss function.
\end{enumerate}
Let us consider a loss function that can derive an \edl classifier.

\subsection{Learning Dirichlet Distributions (\desdirichlet)}

\edl classifiers employ a loss function comprising two components: one component seeks to minimize prediction errors, while the other component targets the count of generated evidence pieces for each class. This approach facilitates the classifier in learning to respond with \textit{I do not know} when confronted with ambiguous data points.

To improve accuracy, we can establish a loss function and calculate its Bayes risk concerning the class predictor. A possibility is to choose the sum of squares loss $||{\bf y}_i - \bm{\pi}_i||_2^2$, hence
\begin{align}
\mathcal{L}_i(\theta) =& \int \underbrace{||\bm{y}_i - \bm{\pi}_i||_2^2}_{\text{SSE loss}} \underbrace{\frac{1}{B(\bm{\alpha}_{i})} \prod_{j=1}^K \pi_{ij}^{\alpha_{ij}-1}}_{\Dir{\bm{\pi_i}|\bm{f}_{\theta}(\bm{x}_i)+\bm{1}}} d \bm{\pi}_i \nonumber \\
=& \sum_{j=1}^K \mathbb{E} \Big [ {y}_{ij}^2 - 2 {y}_{ij} \pi_{ij} + \pi_{ij}^2 \Big ]= \sum_{j=1}^K \Big ( {y}_{ij}^2 - 2 {y}_{ij} \mathbb{E} [ \pi_{ij}  ] + \mathbb{E} [ \pi_{ij}^2 ] \Big )\label{eq:sse}
\end{align}
where ${\bf y}_i$ is a one-hot binary vector encoding the ground-truth class of observation ${\bm x}_i$ with $y_{ij}=1$ and $y_{ik}=0$ for all $k \neq j$, and $\bm{\alpha}_i$ is the parameters of the Dirichlet density on the predictors. Other alternative loss functions have been discussed in \cite{edl}.

\Cref{eq:sse} can be rewritten in the following easily interpretable form, where $\bar{\pi}_{ij} = \mathbb{E} [ \pi_{ij}]$,
\begin{align*}
\mathcal{L}_i(\theta) =&\sum_{j=1}^K (y_{ij} - \mathbb{E} [ \pi_{ij}  ])^2 + \mathrm{Var}(\pi_{ij}) 
= \sum_{j=1}^K \underbrace{(y_{ij} - \bar{\pi}_{ij})^2}_{\mathcal{L}_{ij}^{err}} + \underbrace{\dfrac{\bar{\pi}_{ij}(1-\bar{\pi}_{ij})}{(1 + \sum_{k=1}^K \alpha_{ik})}}_{\mathcal{L}_{ij}^{var}}
\end{align*}
thanks to the fact that 
$\mathbb{E} [ \pi_{ij}^2 ] = \mathbb{E} [ \pi_{ij}]^2 + \mathrm{Var}(\pi_{ij})$.

The loss function aims to minimize the prediction error and the variance of the Dirichlet experiment produced by the neural net for each sample in the training set. This is achieved by breaking down the first and second moments. However, the priority is given to data fit over variance estimation, as indicated by Propositions \ref{prop:propone}, \ref{prop:proptwo}, and \ref{prop:propthree} (proofs are in the supplementary material of \cite{edl}).

\begin{proposition}
For any $\alpha_{ij} \geq 1$, the inequality $\mathcal{L}_{ij}^{var} < \mathcal{L}_{ij}^{err}$ is satisfied.
\label{prop:propone}
\end{proposition}

The subsequent stage is to assess  Equation \ref{eq:sse}'s capacity to align with the data. This assurance is accomplished through the following proposition (the proof is in the supplementary material of \cite{edl}).

\begin{proposition}
\it For a given sample $i$ with the correct label $j$, $L_i^{err}$ decreases when new evidence is added to $\alpha_{ij}$ and increases when evidence is removed from $\alpha_{ij}$.
\label{prop:proptwo}
\end{proposition}

Achieving a strong data fit involves generating a substantial amount of evidence for all classes, as long as the ground-truth class receives the majority of this evidence. However, for capturing epistemic uncertainty, the model should also acquire an understanding of variances that align with the characteristics of the observations. Consequently, when the model is highly confident in its predictions, it should generate more evidence, reflecting this certainty in the outcome. Conversely, the model should refrain from generating evidence for observations it cannot effectively explain. 

Our subsequent proposition (the proof is in the supplementary material of \cite{edl}) furnishes a guarantee for the desired behavior pattern. This pattern, recognized in the uncertainty modeling literature as \textit{learned loss attenuation} \cite{kendall2017what}, is crucial.

\begin{proposition}
For a given sample $i$ with the correct class label $j$, $L_i^{err}$ decreases when some evidence is removed from the biggest Dirichlet parameter $\alpha_{il}$ such that $l \neq j$.
\label{prop:propthree}
\end{proposition}

From the preceding propositions, it follows that when neural networks use the loss function defined in Equation \ref{eq:sse}, they aim to enhance evidence for accurate class labels, thereby reducing misclassifications and decreasing prediction variance on the training data, as long as the new evidence improves data fit.

Regarding the second part of the loss function, \edl encourages the reduction of evidence for incorrect classes and reduces the total evidence count when evidence for the correct class cannot be produced.
For a multi-epoch training procedure, it becomes
\begin{align}
\label{eq:lossfull}
\mathcal{L}(\theta) = \sum_{i=1}^N \mathcal{L}_i(\theta) + \lambda_t \sum_{i=1}^N  \KL{\Dir{\bm{\pi}_i| \bm{\tilde{\alpha}_i}}}{\Dir{\bm{\pi}_i | \bm{1}}}.
\end{align}
In \Cref{eq:lossfull} 
$t$ is the current training epoch,
$\lambda_t = \min(1.0, t / 10 ) \in [0,1]$ is an annealing coefficient,  
$\bm{1}=[1, \ldots, 1]$, 
and $\bm{\tilde{\alpha}}_i = \mathbf{y}_i + (1-\mathbf{y}_i) \odot
\bm{\alpha}_{i}$ is the Dirichlet parameters after removal of the non-misleading evidence from predicted parameters $\bm{\alpha}_{i}$ for sample $i$.\footnote{$\odot$ is the the Hadamard (element-wise) product.}

The KL divergence term in the loss can  be calculated as
\begin{equation}
\label{eq:klreg}
\begin{split}
\KL{& \Dir{\bm{\pi}_i |\bm{\tilde{\alpha}}_i}}{\Dir{\bm{\pi}_i |\bm{1}}} \\ 
&= \log \Bigg ( \frac{\Gamma(\sum_{k=1}^K \tilde \alpha_{ik})}{\Gamma(K) \prod_{k=1}^K \Gamma(\tilde \alpha_{ik}) } \Bigg )+ \sum_{k=1}^K (\tilde \alpha_{ik}-1) \Bigg [\psi(\tilde \alpha_{ik})-\psi \Big (\sum_{j=1}^K \tilde \alpha_{ij} \Big ) \Bigg ],
\end{split}
\end{equation}
where $\Gamma(\cdot)$ is the \textit{gamma} function and $\psi(\cdot)$ is the {\it digamma} function.
To avoid rapid convergence of the neural network toward a uniform distribution for misclassified samples, we systematically and progressively amplify the impact of the KL divergence on the loss using the annealing coefficient.
This enables the network to explore parameter space, preventing premature convergence and improving the chances of correctly classifying mislabeled samples in subsequent epochs.

Empirical evidence \cite{edl} indicates that when a network model is trained via the \edl framework, \edl can maintain the accuracy of the default network model (with softmax). More importantly, \edl provides a measure of uncertainty calibrated to misclassifications and can identify out-of-distributions and adversarial samples.

\subsection{Knowledge Transfer from Pre-trained Models (\destransfer)}
\label{sec:pretraining}
In the previous sections, we presented \edl as a stand-alone training strategy for neural networks. It can be used to train a neural network from scratch for a classification task. 

We argue that the \edl loss function can be used to transfer knowledge \--- \ie fine-tune \--- from a pre-trained deep classifier to enhance its predictive uncertainty and turn it into an evidential classifier (\destransfer).
This is not the case for most of the existing approaches for uncertainty quantification because they require architectural changes such as additional dropout layers and modelling of weight distributions at each layer.

Applicability of uncertainty quantification approaches like \edl to pre-trained models is essential for their usability in realistic problem settings and adoption by the larger community.
Pre-trained models are frequently used in industry and academia, especially in a transfer learning setting, where a network is once trained using a large amount of data on a generic classification task and then tuned for other tasks using a smaller amount of specialized datasets. 

Pre-trained models most likely discovered patterns in the data: \eg the presence of a prominent circular pattern in an image from the MNIST dataset can be indicative of the digit zero. 
By re-training such models using \edl's loss, the regularizing term described in \Cref{eq:klreg} will penalize divergences from the {\it ``I do not know''} state that do not contribute to the data fit.

To show that \edl satisfies \destransfer, let us consider pre-trained deep learning models from \textit{pytorchcv} model library\footnote{\url{https://pypi.org/project/pytorchcv}, on 17th Aug 2023.} for CIFAR10, CIFAR100, and  ImageNet-1K datasets.
We used Resnet56 pre-trained on CIFAR10 and CIFAR100 datasets for $500$ and $360$ epochs, respectively. For ImageNet, we used Resnet50 pre-trained for $486$ epochs.
Test accuracies of these models are $0.9548$, $0.7512$, and $0.777$ for these datasets, respectively. 
We tuned these models for $10$ epochs with their original training set and procedure but with the \edl loss. We used Adam optimizer with a learning rate of $\num{1e-5}$. 
While generating evidence from deeper networks, we found that clamped exponent function $\exp(\min(x, 10))$ works well. 
We also added an additional term $x - bg(x)$, where $bg(\cdot)$ is an identity function with blocked gradients, \ie $bg(x)=x$ and $\nabla bg(x)=0$. The resulting evidence function has the form
\begin{equation}
f(x) = \exp(\min(x, 10)) + (x - bg(x)),
\label{eq:clamping}
\end{equation}
which has a non-zero gradient for all valid values of $x$ despite the clamping. 
The exponential function is clamped for numerical stability, particularly during initial epochs. Logits can exhibit significant fluctuations and large values, leading to numerical instability (\eg \texttt{NaN} errors). By clamping logits to a small value (\eg 10), we mitigate this instability while still allowing the network to produce substantial evidence (\eg $\exp(10) > 22 \cdot 10^3$). However, clamping results in zero gradients, adversely affecting training, especially in the early stages. To maintain gradients as if no clamping were applied, we employ a straight-through gradient trick in \cref{eq:clamping}. The term $(x - bg(x))$ does not alter the exponent function's value (evaluating to zero), ensuring consistent gradient flow during backpropagation.

Table~\ref{table: tune} presents our test results for pre-trained models and after these models are tuned using \edl loss and their original training data.
The table demonstrates the test accuracy values of models. It also presents the normalized area under the curve (AUC) for the empirical entropy CDF of model predictions. We normalized the area under these curves using their maximum value, corresponding to the logarithm of the number of classes: indeed, entropy is maximum and equal to $\log K$, equivalent to a uniform distribution over  $K$ classes.
The normalized AUC is between 0 and 1;  we expect it to be lower when the model fails in its predictions or encounters out-of-distribution (OoD) samples.

\begin{table}
\caption{Test set results for pre-trained models and after they are tuned using \edl loss and training data for CIFAR10, CIFAR100, CIFAR110 (which combines CIFAR10 and CIFAR100 datasets), and ImageNet datasets. \label{table: tune}}
\centering
{\renewcommand{\arraystretch}{1.3}
\begin{tabular}{ l l }
\toprule
\textbf{Dataset} & \hspace{2.0in}\textbf{Normalised Entropy AUC} \\
        & \begin{tabular}{p{0.8in} p{0.8in} p{0.5in} p{0.7in} p{0.4in}}
~ & Accuracy & Correct & Incorrect & OoD\\
\end{tabular}\\
\midrule
CIFAR10 & \begin{tabular}{p{0.8in} p{0.8in} p{0.5in} p{0.7in} p{0.4in}}
Pretrained & 0.952 & 0.974 &	0.776 &	0.759\\
\edl-tuned  &0.948 & 0.953 & 0.375 & 0.425\\ 
\end{tabular} \\
\hline
CIFAR100 & \begin{tabular}{p{0.8in} p{0.8in} p{0.5in} p{0.7in} p{0.4in}}
Pretrained & 0.751 & 0.849 & 0.636 & 0.55\\ 
\edl-tuned  & 0.744 & 0.801 & 0.195 & 0.104\\ 
\end{tabular} \\
\hline
IMAGENET & \begin{tabular}{p{0.8in} p{0.8in} p{0.5in} p{0.7in} p{0.4in}}
Pretrained &  0.777 & 0.815 & 0.601 & 0.475 \\ 
\edl-tuned  & 0.775 & 0.819 & 0.357 & 0.302 \\ 
\end{tabular} \\
\hline
CIFAR110 & \begin{tabular}{p{0.8in} p{0.8in} p{0.5in} p{0.7in} p{0.4in}}
Pretrained & 0.624 & 0.747 & 0.679 & - \\ 
\edl-tuned  & 0.735 & 0.825 & 0.401 & - \\ 
\end{tabular} \\
\bottomrule
\end{tabular}
}
\end{table} 

\Cref{table: tune} indicates that \edl can significantly improve the model's predictions in terms of uncertainty without sacrificing its test accuracy. The models tuned using \edl loss provide much higher uncertainty when their predictions are incorrect, or they are given OoD samples. The higher uncertainty means high entropy and a smaller area under the entropy CDF. 
We used the CIFAR100 test set as OoD samples for CIFAR10 and ImageNet\footnote{We resized CIFAR100 images to the size of the images in the ImageNet dataset.} classifiers. 
Let us note that CIFAR10 and CIFAR100 have a disjoint set of labels even though they look similar.

\subsection{Fusion of Evidential Classifiers (\descompositionality)}
\label{sec:fusion}

To support effective decision-making in the real world, a necessary characteristic is the possibility \--- for the classification system \--- \textbf{to manifest some ability of compositionality} (\descompositionality), \ie it is desirable to be able to fuse the predictions of two classification systems on different sets of categories. 
In this section, we formalize how predictions of two \edl classifiers can be fused on different sets of categories.

Indeed, since its original conception in \cite{edl}, \edl has already been used extensively in producing various deep learning models.
Bauer \etal proposed to use \edl to build spatially coherent inverse sensor models for self-driving cars~\cite{bauer2019deep}.
Ghesu \etal used \edl to quantify and leverage the classification uncertainty for chest radiography assessments \cite{ghesu2019quantifying}. 
They reported that the existing dataset of chest radiography has large label noise due to automatic labelling from x-ray reports: \edl's uncertainty prediction can be used as a  filter.
Sahin \etal proposed to use \edl's predictive uncertainty estimates to improve search results for medical specialty search ~\cite{csahin2019medspecsearch}.
Hu and Khan used the predictive uncertainty of \edl for reliable text classification and out-of-distribution detection~\cite{hu2021uncertainty}.
Bao \etal used \edl for open set action recognition ~\cite{bao2021evidential}.
Hemmer \etal used it for active learning of image classification~\cite{hemmer2022deal}.
Wang \etal employed \edl to calculate the uncertainty of robots' decisions in an active learning setting, effectively reducing the amount of human feedback needed for a robot to learn a given task \cite{wang2020maximizing}. 
Amini \etal extended \edl for regression tasks~\cite{amini2020}. 
Soleimany \etal used it for guided molecular property prediction and discovery ~\cite{soleimany2021evidential}.

Let us define two disjoint sets of categories $X$ and $Y$, \ie $X  \cap  Y = \varnothing$, and $|A|=n$ and $|B|=m$.
For example, $A$ and $B$ could represent the categories in CIFAR10 and CIFAR100 datasets, respectively. 
The categories used as labels in these datasets are disjoint even though the images look similar. 

Let us have two \edl classifiers $C_{A}(\cdot)$ and $C_{B}(\cdot)$ trained on classification tasks defined over $A$ and $B$, respectively. For any sample $\bm{x}$, these classifiers predict a Dirichlet distribution over the corresponding sets of categories, \ie $C_{A}(\bm{x})=\Dir{\bm{\pi_{A}}|\bm{\alpha_{A}}}$ and $C_{B}(\bm{x})=\Dir{\bm{\pi_{B}}|\bm{\alpha_{B}}}$, where $\bm{\alpha_A} = \alpha_{A1},\ldots,\alpha_{An}$ and $\bm{\alpha_B} = \alpha_{B1},\ldots,\alpha_{Bm}$.

Using the aggregation and neutrality properties of Dirichlet distributions~\cite{frigyik2010introduction}, we can merge the evidence in the predicted Dirichlet distributions to have a prediction over the union of categories $A \cup B$ as 
$$C_{A \cup B}(\bm{x}) = \Dir{\bm{\pi_{A \cup B}}| \alpha_{A1},\ldots,\alpha_{An}, \alpha_{B1},\ldots,\alpha_{Bm}}.$$

\begin{figure}[t!]
$\begin{array}{cc}
\hspace{-0.25in} \includegraphics[width=7cm]{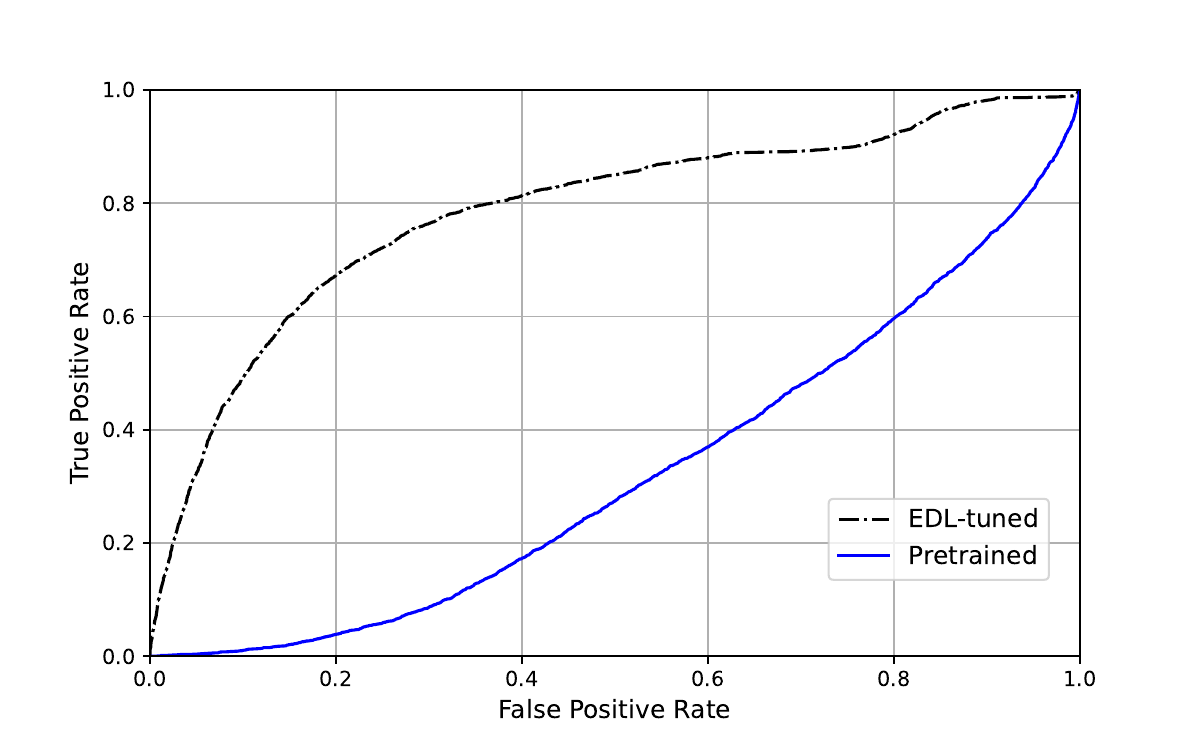} & 
\includegraphics[width=7cm]{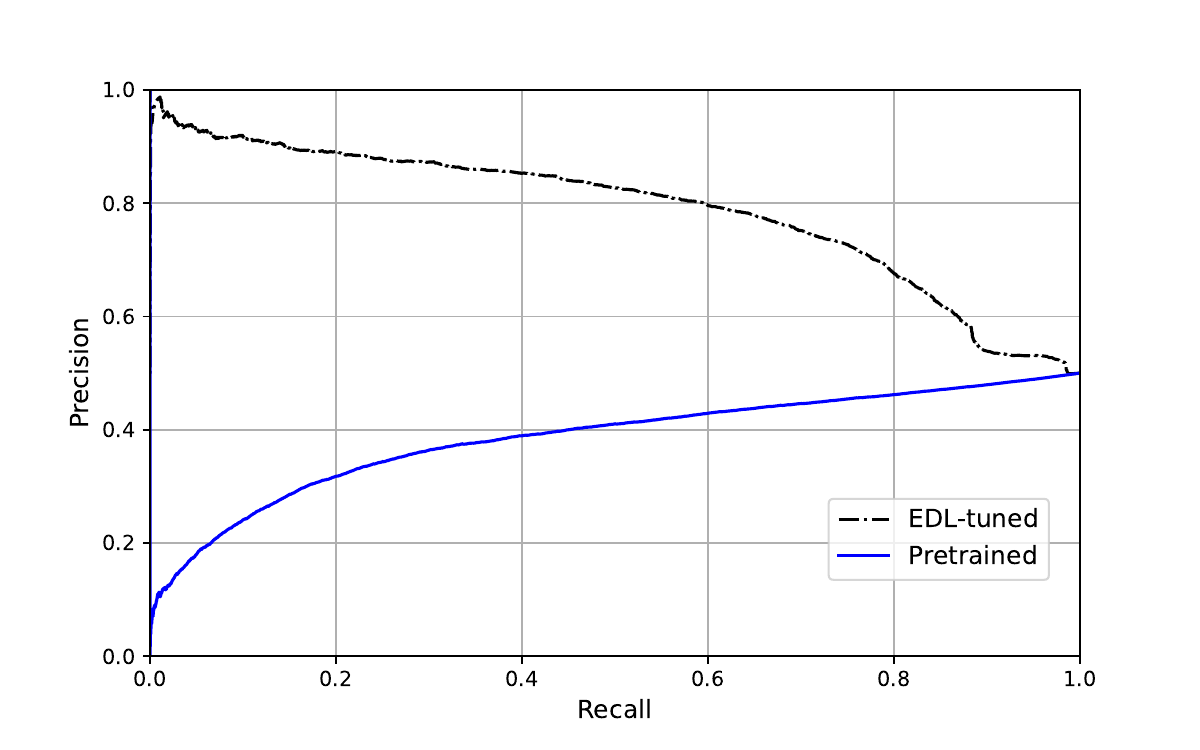}
\end{array}$
\caption{\label{fig:fusion_cifar} ROC (left) and PRC (right) for the fusion of pre-trained and \edl-tuned classifiers. We used the entropy of predictions as uncertainty scores, and the area under the curves indicates how useful uncertainties are while discriminating between correct and incorrect predictions.}
\end{figure}

In \Cref{table: tune}, CIFAR110 combines CIFAR10 and CIFAR100 datasets.
We combine CIFAR10 and CIFAR100 classifiers to classify CIFAR110 samples by concatenating the output logits of the corresponding models.
The fused pre-trained classifiers using this approach could only achieve 62.4\% accuracy on the combined test sets of CIFAR10 and CIFAR100. 
We also tried calibration by temperature scaling of logits before fusing classifiers, but it did not improve the performance. 

However, when we fuse the \edl-tuned pre-trained models using the same approach, we had 73.5\% accuracy. 
This indicates that \edl allows models to predict Dirichlet distributions that can be effectively combined by aggregating evidence for different categories.
Furthermore, tuning with \edl loss gave models much better predictive uncertainties.

\Cref{table: tune} indicates that fine-tuning with \edl loss allows us to have better uncertainty quantification for predictions and easily fuse classifiers trained on data from a disjoint set of classes.

To understand the predictive uncertainties of the fused classifiers, we showed ROC and Precision and Recall Curves (PRC) for the pre-trained and \edl-tuned models in \Cref{fig:fusion_cifar}. 
While calculating these curves, we used uncertainty as scores to estimate how useful these uncertainties are to discriminate between correct and incorrect predictions.

For ROC, we found that the AUC is $0.323$ and $0.782$ for pre-trained and \edl-tuned models, respectively. 
Also, for PRC, the AUC becomes $0.382$ and $0.786$ for these models, respectively. 
The area under these curves becomes larger as the predictive uncertainties get more useful in discriminating between correct and incorrect predictions.
These results are aligned with our previous observations regarding the \edl's ability to turn pre-trained models into models with better predictive uncertainties.

\section{Risk-aware Classification under Uncertainty}\label{sec:decision}
After showing, in \Cref{sec:edl_learning}, that \edl does indeed satisfy the three desiderata discussed in \Cref{sec:motivation_full_section} \--- and it does it so efficiently, thus without requiring, for instance, external datasets \--- we now turn our attention to derive \edl variations embedding risk awareness.

Let's revisit pignistic probabilities \cite{smets_DecisionmakingTBM_05}, introduced in decision theory to represent how likely a rational agent is to choose a specific option in decision-making (as discussed in \Cref{sec:motivation_full_section}). Pignistic probabilities, denoted as $\bm{p}=[p_1,\ldots, p_K]$, are mathematically equivalent to the Shapley value in game theory~\cite{dubois2008definition}. They encapsulate uncertainty when faced with $K$ options, reflecting both uncertainty related to $\bm{\pi}$ and the associated risk for each option.
Hence, when assessing the expected risk of assigning a category as the label for a sample, pignistic probabilities ($\bm{p}$) are used instead of categorical probabilities ($\bm{\pi}$) to consider the risk of misclassification properly.

Concerning a classification decision for $\bm{x}$, the following Dirichlet distribution can represent the associated pignistic probabilities:
\begin{equation}
    q_\theta(\bm{p}| \bm{x}) = \Dir{\bm{p} | c_{\theta}({\bm x}) + \gamma_\Theta(\bm{x})},
\end{equation}
In this formulation, we replace the uniform prior $\bm{1}$ with a prior that is specific to each sample, denoted as $\gamma_{\Theta}(\bm{x})$. It is a per-sample redistribution of uniform prior counts over $K$ categories,
\begin{equation}
    \label{eq:pignisticprior}
    \bm{\gamma}_{\Theta}(\bm{x}) = K \text{softmax}(\bm{W} g_{\theta}(\bm{x}) + \bm{b}),
\end{equation} 
and represents the prior count for the pignistic probabilities.
We employ the symbol $\Theta$ to denote trainable variables, including $\bm{W}$ and $\bm{b}$. These variables are utilized to compute $\gamma_{\Theta}(\bm{x})$ based on $g_{\theta}(\bm{x})$, which signifies the input to the \textit{logits layer} of the neural network.

\begin{figure}
    \centering
    \includegraphics[width=\linewidth]{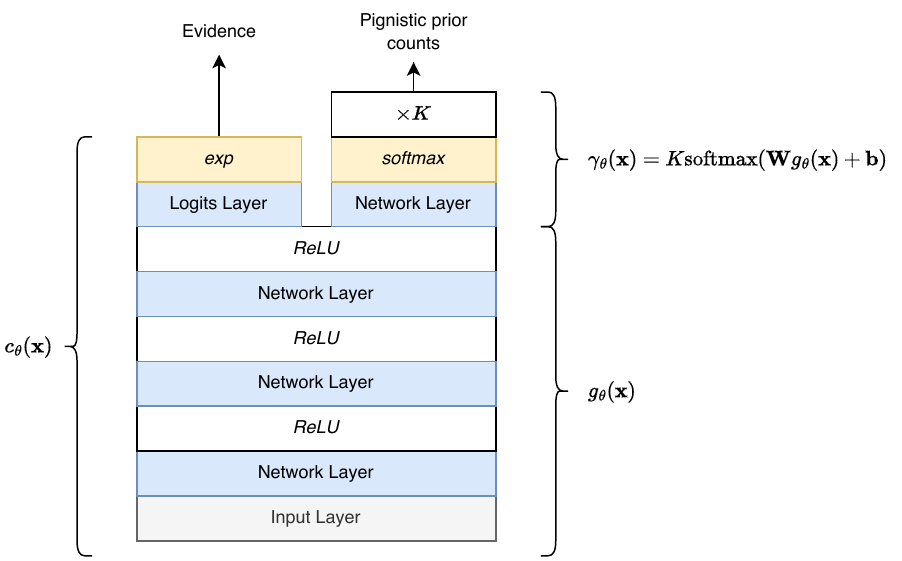}
    \caption{Proposed architecture for risk-aware classification.}
    \label{fig:architecture-risk-aware}
\end{figure}

We clarify the architecture of our proposed network in \Cref{fig:architecture-risk-aware}. The network comprises two heads: one for predicting pseudo-counts and the other for predicting pignistic priors. These heads share most of their parameters, essentially forming the backbone of the network. This configuration resembles multi-task learning, where two supporting tasks are learned using shared parameters. In this context, proficiency in uncertainty quantification becomes paramount. The influence of the pignistic prior \--- whose parameters are denoted as  \(W\) and \(b\) \--- will rapidly diminish if the network predicts a large number of misleading evidence, where the posterior is the sum of the prior and the evidence.

It is important to emphasize that the summation of prior counts for the pignistic probabilities, denoted as $\sum_i \gamma_{\Theta i}(\bm{x})$, equals $K$. In simple terms, the total prior counts for pignistic probabilities ($\bm{p}$) match those of categorical probabilities ($\bm{\pi}$). This equivalence ensures that the network's predicted evidence for categorical probabilities (\ie $c_{\theta}(\bm{x})$) aligns with the prior counts for pignistic probabilities.
This modeling approach permits variations in the risk of misclassification between samples, influenced by factors like the true category. By employing this approach, the neural network learns to assign lower risk to specific categories and subsequently increases their prior counts via $\gamma_{\Theta}(\cdot)$.

Using the pignistic probabilities, we determine the mean misclassification risk for $\bm{x}$ concerning its true category $y$ with
\begin{equation*}
\text{risk}(\bm{x}) = \sum_{i=1}^K R_{yi} ~p_i    
\end{equation*}
where $R_{yi}$ denotes the misclassification cost for classifying as $i$ a sample that belongs to the class $y$.
To compute the expected risk, we can integrate the pignistic probabilities expressed as Dirichlet distributions with $\bm{\alpha}$ such that  $\alpha_i=c_{\theta i}(\bm{x}) + \gamma_{\Theta i}(\bm{x})$:
\begin{align}\label{eq:exp_risk}
\hspace{-0.15in}
\mathbb{E}[\text{risk}(\bm{x})]
  &= \mathbb{E}_{q_\theta(\bm{p}|\bm{x})}\big[\mathbb{E}_{i \sim \bm{p}}[R_{yi}]\big]  
  = \frac{\sum_{i=1}^{K} R_{yi} (c_{\theta i}(\bm{x}) + \gamma_{\Theta i}(\bm{x}))}{K + \sum_{j=1}^{K} c_{\theta j}(\bm{x})},
\end{align}
where $c_{\theta i}(\bm{x})$ and $\gamma_{\Theta i}(\bm{x})$ corresponds to the $i^{th}$ element of the vector outputs from these functions.

In the following, we introduce three approaches for embedding risk awareness in \edl: {\textbf{riskEDL}, {\textbf{EDL(p)}, and {\textbf{EDL(pg)}. With reference to \Cref{fig:architecture-risk-aware}, this implies different training methodologies for the network heads. In the \textbf{EDL(p)} method, the pignistic prior parameters \---  \(W\) and \(b\) \--- are trained after the \edl network for pseudo-counts has been trained and its parameters frozen. 
The rationale behind freezing the network parameters in the \textbf{EDL(p)} method is to circumvent potential degradation of both classification accuracy and uncertainty quantification. This is particularly important in scenarios where there is a high asymmetric cost of misclassification. For instance, the network may learn to frequently choose low-cost classes to avoid risk, even though these classes are more likely to be incorrect.
In the \textbf{riskEDL} method, both heads are trained simultaneously without freezing any parameters. We will also consider a case where we do not use supervised learning but rather with trial-and-error, \textbf{EDL(pg)}.

\subsection{{\textbf{riskEDL}}: Combining \edl loss with a risk-awareness related loss}

A way to train agents for risk awareness is to include risk minimization within the training loss of the \edl classifiers and train the network to learn uncertainty quantification and risk minimization simultaneously. 
We name this first approach \textbf{riskEDL} since it combines the \edl loss with a risk-awareness related loss~\cite{DBLP:conf/wacv/SensoySJAR21}. 

In this method, we avoid freezing any weights in the network. This leads to a multi-objective optimization process that minimizes risk and enables classification with an awareness of uncertainty. While doing so, the denominator in Equation ~\ref{eq:exp_risk} is removed since it encourages the generation of more evidence for less risky categories and degenerates uncertainty quantification. 

More precisely, we add the following loss 
$$
\kappa \sum_{i=1}^{K} R_{yi} (c_{\theta i}(\bm{x}) + \gamma_{\Theta i}(\bm{x})),
$$
to the \edl loss introduced in \Cref{sec:edl_learning} during training, where $\kappa$ represents the weight of this additional loss, set to $0.01$ for any of the empirical results of this paper.

\subsection{{\textbf{EDL(p)}}: Minimizing the Risk of Misclassifications}
In this approach, we will be learning only the $\bm{W}$ and $\bm{b}$ parameters of $\gamma_{\Theta}(\bm{x})$ while keeping the rest of the network unchanged. 
To minimize the risk of misclassification, we can directly minimize the expected risk in Equation \ref{eq:exp_risk} as an additional loss after training and freezing the weights of the EDL network.
We name this approach \textbf{EDL(p)}, where $p$ refers to the pignistic probabilities.

\subsection{{\textbf{EDL(pg)}}: Policy Gradient for Learning Pignistic Priors}
In the previous sections, we assumed that the risk-awareness is learned using supervised learning with an explicit risk matrix. 
However, the risk matrix may not be available for the agent ahead, and the learning may be performed through trial-and-error as in many real-life settings.
This problem can be addressed by an approach based on the policy gradient method. 

In our context, the classification policy of an agent is defined as the expected probability that it picks category $i$ as the label of a sample $\bm{x}$. It is computed using
\begin{align}\label{eq:policy}
P(i|\bm{x}) = \int  p_i q_\theta(\bm{p}| \bm{x}) d \bm{p}=  \frac{c_{\theta i}(\bm{x}) + \gamma_{\Theta i}(\bm{x})}{K + \sum_{j=1}^{K} c_{\theta j}(\bm{x})}.
\end{align}

\begin{myalgorithm} 
\caption{Policy Gradient for Learning Pignistic Priors \label{alg:risk(pg)}}
\begin{algorithmic}[1]
\Require $\mathcal D$: dataset, $\lambda$: learning rate, $\Theta = \{\bm{W}, \bm{b}\}$: trainable parameters of $\gamma_{\Theta}(\bm{x})$, $\bm{R}$: risk matrix%
\State Randomly Initialize $\Theta$
\ForEach {$(\bm{x},y) \in \mathcal D $}
\State $i \sim P(i|\bm{x})$ \Comment{$P(i|\bm{x})$ is defined in Equation \ref{eq:policy}~~~~~}
\State $\Theta \gets \Theta - \lambda  R_{yi} \nabla_{\Theta} \log P(i|\bm{x})$
\EndFor 
\end{algorithmic}
\end{myalgorithm}

Given the classification policy of the agent, we want to update parameters $\Theta$ to minimize the expected risk of decision-making,
$$\sum_{(\bm{x}, y)} \mathbb{E}_{i \sim P(i|\bm{x})}  \Big [ R_{yi} \Big ],$$ 
where $(\bm{x}, y)$ tuple represents each sample $\bm{x}$ and its true label, and $i$ represents the predicted label for $\bm{x}$. Then, the update rule of the parameters for gradient descent, for each sample $\bm{x}$ with truth label $y$, can be written as
\begin{align}\label{eq:pg}
\theta = \theta - \lambda \nabla_{\theta} \mathbb{E}_{i \sim P(i|\bm{x})} \Big [ R_{yi} \Big ],
\end{align}
where $\lambda$ is the learning rate. The gradient of the expectation can be rewritten using log-trick~\cite{sutton2018reinforcement} as follows:
\begin{align}\label{eq:pg}
\nabla_{\theta} \mathbb{E}_{i \sim P(i|\bm{x})} \Big [ R_{yi} \Big ] = \mathbb{E}_{i \sim P(i|\bm{x})} \Big [ R_{yi} \nabla_{\theta} \log P(i|\bm{x}) \Big ],
\end{align}
which can be approximated using Monte Carlo sampling \cite{murphy2012machine}. The pseudo-code of the resulting algorithm is shown in Algorithm~\ref{alg:risk(pg)}, where the agent makes only one decision for each sample at each epoch, and these decisions are independent. 
Hence, our problem is similar to the contextual bandit problem, where the episode length is just one. 

Let us note that our approach to risk-awareness is tightly coupled with the correct uncertainty estimation for prediction. 
That is, if a false prediction has very high uncertainty, the agent's decision will mostly depend upon the pignistic prior, which reduces the cost of misclassification for the agent. On the other hand, if the false prediction has very low uncertainty, the effect of the pignistic prior on the decision would be small, which may lead to a prediction without risk-awareness. 

We refer to this approach as \textbf{EDL(pg)} in the rest of the paper, and it can be considered as an adaptation of the well-known REINFORCE algorithm~\cite{sutton2018reinforcement} for our setting.

\section{Evaluation}
\label{sec:evaluation}
In this section, we assess the applicability of our approaches for minimizing the expected cost of misclassification in classification decisions by agents. While \edl is a versatile method that can be applied to any deep neural network for classification, our evaluation focuses on the MNIST 
and CIFAR10 
datasets.
To ensure comparability, we adopt the experimental framework previously explored by Louizos et al. \cite{Louizos17}. We choose the LeNet~\cite{lecun1998gradient} architecture for our neural network, as it has been widely used as a benchmark for assessing uncertainty quantification in deep neural networks~\cite{rahaman2021uncertainty, nilsen2022epistemic, Louizos17}. The neural network architecture is the standard LeNet with ReLU. All experiments are implemented using Tensorflow~\cite{abadi2016tensorflow}, and training utilizes the Adam optimizer~\cite{kingma2015method} with default settings.
For the MNIST dataset, we train the LeNet architecture~\cite{lecun1998gradient} with $20$ and $50$ filters of size $5 \times 5$ in the first and second convolutional layers, respectively, and a fully connected layer consisting of 500 hidden units. In the case of CIFAR10, we employ the larger LeNet version, equipped with $192$ filters at each convolutional layer and $1000$ hidden units for the fully connected layers. 
The MNIST dataset comprises $60,000$ training samples and $10,000$ test samples, whereas the CIFAR10 dataset consists of $50,000$ training samples and $10,000$ test samples. Our models were trained over $50$ epochs, and evaluation was performed on the test samples.

\begin{figure*}
  \resizebox{\textwidth}{!}{%
    \begin{tabular}{cc}
      \includegraphics[width=3.5cm]{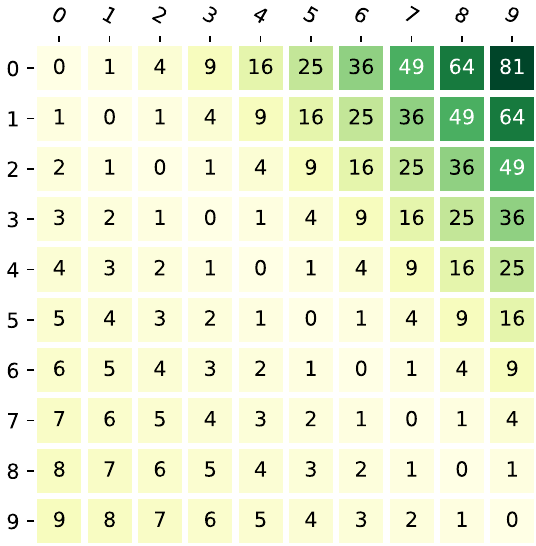} &
      \includegraphics[width=4.5cm]{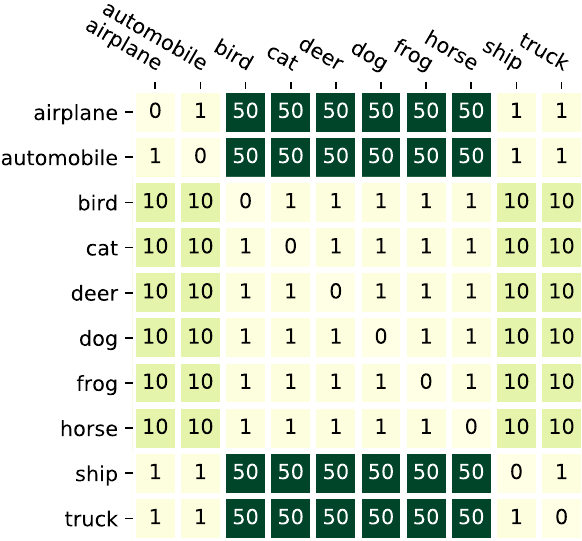}
    \end{tabular}
  }
\vspace{-0.12in}
  \caption{\label{fig:risk_mat}Risk matrices.}
\end{figure*}

We compare our approaches with: (a) \textit{standard} learning, which uses the \textit{cross-entropy} loss (\textbf{L2}); (b) a \textit{cost-sensitive} learning approach (\textbf{CS-L2}) by Galdran \textit{et al.} 
\cite{galdran2020cost}, which regularizes the \textit{standard} learning using the misclassification cost defined in the risk matrix; and (c) \textbf{EDL} as introduced in 
Section~\ref{sec:edl_learning}.
In our risk-aware \edl classifiers, namely \textbf{riskEDL}, \textbf{EDL(p)}, and \textbf{EDL(pg)}, we employ the expectation of the predicted Dirichlet distribution for pignistic probabilities, denoted as $q_\theta(\bm{p}| \bm{x})$, as the predictive distribution.
All of above models utilize the identical LeNet architecture; however, they employ different loss functions.

In our experiments, we employed two distinct risk matrices, illustrated in Figure~\ref{fig:risk_mat}. For the MNIST dataset, the risk matrix is such that 
\[
R[i,j] = \left\{\begin{array}{l l}
    (i-j)^2 & \text{if~} j > i;\\
    (i-j) & \text{otherwise},
\end{array}\right.
\]
where $i$ is the index of the ground-truth category, and $j$ indexes the predicted one.
For MNIST, misclassifing an image for a digit with greater value incurs a greater cost compared to undervaluing it.
For the CIFAR10 dataset, $R[i,j]=1$ if both $i$ and $j$ belong to the same group, \ie either \textit{animals} or \textit{vehicle}. If, instead, we misclassify an animal, the cost is $10$; otherwise, is $50$ \--- \cf Figure~\ref{fig:risk_mat} (right).

The risk matrix for CIFAR10 is intuitive, especially for applications such as self-driving cars, where the cost of misclassifying an animal as a vehicle may have a significantly different cost than the opposite. 
Similarly, this risk matrix defined for MNIST may have several real-life applications in autonomous agents and multi-agent systems. For instance, an agent in seal-bidding auctions may desire to classify the auctioned item into several ordinal categories based on its value (\eg ordinary, luxury, antique) and place the bid based on this classification.
Classifying the item into a more valuable category may lead to a higher bid.
In this case, the agent may have a similar risk matrix to avoid significantly overestimating the value of the auctioned item, where the underestimation of the value may lead to losing the auction by placing a lower bid instead of losing a significant amount of money by overbidding.

\begin{figure}
\begin{floatrow}
\capbtabbox{%
\begin{tabular}{ccc} \toprule
  \textbf{Model} & \textbf{MNIST}& \textbf{CIFAR10} \\ \midrule
  \textbf{L2} & \textbf{99.4} & 72.1  \\
  \textbf{CS-L2} & 99.35 & 72.3 \\
  \textbf{EDL} & 99.3  & 72.8 \\
  \textbf{riskEDL} & 99.3 & 73.0 \\
  \textbf{EDL(p)} & 99.32 & 72.4 \\
  \textbf{EDL(pg)} & 99.28 & \textbf{73.6} \\
  \bottomrule
  \end{tabular}
}{%
  \caption{Test set accuracy\label{tab:acc}}%
}
\capbtabbox{%
 \begin{tabular}{ccc}
        \toprule
  \textbf{Model} & \textbf{MNIST} & \textbf{CIFAR10} \\
  \midrule
  \textbf{L2} &  9.23 & 11.39 \\
  \textbf{CS-L2} & 7.47 & 4.92 \\ 
  \textbf{EDL} & 7.91 & 10.37 \\
  \textbf{riskEDL} & 3.37 & 3.07 \\
  \textbf{EDL(p)} & \textbf{2.92} &  \textbf{2.58} \\
  \textbf{EDL(pg)} & 3.02 & 2.85 \\
  \bottomrule
  \end{tabular}
}{%
  \caption{Misclassification cost for test set. \label{tab:cost}}%
}
\end{floatrow}
\end{figure}

Tables~\ref{tab:acc} and~\ref{tab:cost} showcase our test results for average misclassification cost and accuracy. They reveal that cost-sensitive training surpasses standard and evidential deep learning, yielding lower misclassification costs. This is unsurprising, given that these methods do not consider the risk matrix during training.

Notably, \edl utilizing pignistic probabilities achieves significantly reduced misclassification costs compared to cost-sensitive training, while maintaining similar or even improved accuracy on the test samples across all datasets. In essence, the proposed risk-aware \edl approaches reduce the cost of misclassification by at least 55\% for MNIST and 38\% for CIFAR10 compared to cost-sensitive training. Moreover, the reduction in misclassification cost becomes even more substantial for \textbf{EDL(p)} and \textbf{EDL(pg)}.

\begin{figure}
    \centering
    \begin{subfigure}{0.45\textwidth}
        \centering
        \includegraphics[width=\linewidth]{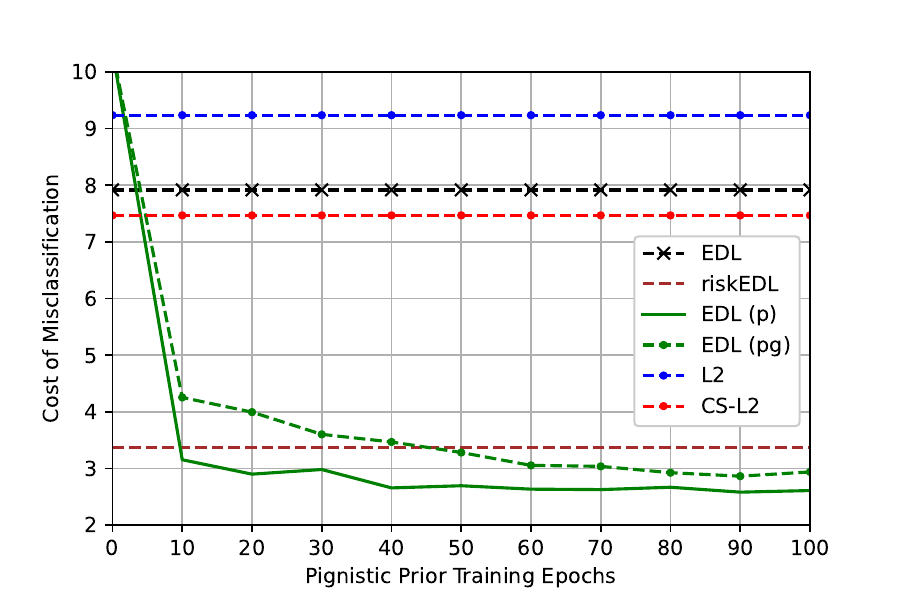}
        \caption{}%
    \end{subfigure}
    \hfill
    \begin{subfigure}{0.45\textwidth}
        \centering
        \includegraphics[width=\linewidth]{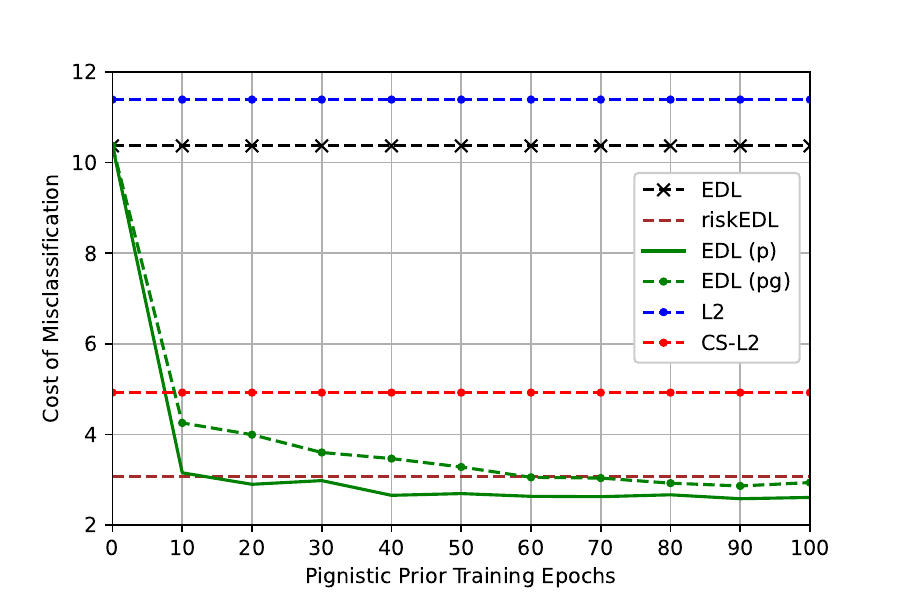}
        \caption{}%
    \end{subfigure}
    \caption{Change in the misclassification cost for MNIST (a) and CIFAR10 (b) test sets over different the pignistic prior training epochs.}
    \label{fig:mnist_cifar10_pignistic}
\end{figure}

Figure \ref{fig:mnist_cifar10_pignistic} presents a detailed overview of the change of misclassification cost for different datasets as we train the network for predicting pignistic priors for \textbf{EDL(p)} and \textbf{EDL(pg)}. The misclassification costs for other approaches are shown as constant for comparison purposes. Let us note that each of \textbf{EDL(p)} and \textbf{EDL(pg)} gets the pre-trained \edl network, freezes its weights, and learns only the additional parameters to predict pignistic priors (\ie $\bm{W}$ and $\bm{b}$).
Let us note that Algorithm~\ref{alg:risk(pg)} represents one epoch of training for the \textbf{EDL(pg)} approach. 
The figures indicate that \textbf{EDL(p)} and \textbf{EDL(pg)} become better than \textbf{riskEDL} after their training for the pignistic prior parameters. Unlike these approaches, \textbf{riskEDL} trains pignistic parameters together with the parameters of the whole network. 

The results show that \textbf{EDL(pg)} becomes better than cost-sensitive training within a few epochs of training. Moreover, it becomes almost as good as \textbf{EDL(p)} as it is trained further.
This indicates that \textbf{EDL(pg)} can learn to make low-cost classification decisions effectively with bandit feedback~\cite{jeunen2019learning} while \textbf{EDL(p)} has access to the complete cost matrix from the beginning. 
Learning from bandit feedback is important in real autonomous systems since this is a realistic setting where the agent can only observe the outcome of its actions, \eg the cost of its current classification decision, instead of the cost of all classification decisions at once for a given sample.

While our evaluations indicate that \textbf{EDL(p)} and \textbf{EDL(pg)} generally outperform other methods, there are important considerations for \textbf{riskEDL}. \textbf{RiskEDL} combines \textbf{EDL} loss with a risk-awareness-related loss, resulting in a multi-objective optimisation problem. This approach can sometimes cause one objective, such as risk or uncertainty, to overshadow the other. Specifically, the model may become more confident in low-risk categories, potentially prioritising these categories at the expense of reducing uncertainty. This prioritisation can occur even if a lower cost could be achieved on the test set by preserving the actual uncertainty in these cases. In contrast, \textbf{EDL(p)} and \textbf{EDL(pg)} primarily factor in risk in cases of low evidence (high uncertainty) by incorporating risk into the prior. We believe this makes these approaches more effective than \textbf{riskEDL} for most risk matrices. However, \textbf{riskEDL} may become more effective when class errors carry equal risks and the risk matrix is binary with a zero diagonal.

\section{Discussion and Related Work}
\label{sec:related}
\subsection{Uncertainty Quantification in Neural Networks}

\edl \cite{Sensoy2018} efficiently \--- \ie without the need for external sources of knowledge, such as additional datasets \--- estimates Dirichlet distributions instead of point probabilities. This approach is not the only one in this direction, as indicated by recent surveys \cite{hullermeier_AleatoricEpistemicUncertainty_21,abdar_reviewuncertaintyquantification_21,cerutti_EvidentialReasoningLearning_22a}. In the following discussion, we will consider some other approaches that could potentially be used for risk-aware classification. However, as of the time of writing, there is no published evidence suggesting their efficient fulfillment of all the requirements we discussed in \Cref{sec:motivation_full_section}. While many of the following approaches can satisfy \desdirichlet, which involves interpreting the output of the classification system as a Dirichlet distribution, we are not aware of any work demonstrating their ability to satisfy \destransfer\xspace (transfer knowledge) or \descompositionality\xspace (compositionality). On the contrary, most related work requires such profound architectural changes that complete re-training becomes necessary, thereby undermining \destransfer and severely hampering the satisfiability of \descompositionality.

Regarding uncertainty quantification, Bayesian deep learning~\cite{Louizos17,kendall2017uncertainties,gal2016icml,lakshminarayanan2017simple,pawlowski2017implicit} combines deep neural networks with Bayesian probability theory. This approach systematically captures uncertainty in machine learning models by introducing a prior distribution on their parameters and inferring the posterior distribution using techniques such as Variational Bayes~\cite{blundell2015icml,gal2016icml,sun2019functional}.
To estimate the posterior predictive distribution, sampling methods are typically employed, but this adds computational overhead and noise. These models quantify predictive uncertainty by generating samples from the posterior distributions of model parameters and using them to create a prediction distribution for each network input. However, it is essential to note that while modeling the uncertainty of network parameters is common, it may not always provide accurate estimates of neural network predictive uncertainty~\cite{hafner2018reliable}.

Pragmatic techniques like deep ensembles~\cite{lakshminarayanan2017simple} and Monte Carlo (MC) dropout~\cite{gal2016icml} have been introduced for estimating predictive uncertainties in neural networks.
In deep ensembles, multiple models are trained separately for the same problem, and their predictions are used during inference time to quantify the predictive uncertainty.
Instead of training multiple models in MC dropout, a single model is trained with dropout layers between network layers. During inference, the dropout layers are activated, and the model is queried multiple times for each input to get predictions for the same input. These predictions are used to estimate the predictive uncertainty as in deep ensembles.
These approaches are used in various settings such as uncertainty-aware medical image classification~\cite{abdar2021barf, senousy2021mcua} and disease detection~\cite{abdar2023uncertaintyfusenet}.
While they are easy to implement, they are computationally demanding due to multiple inferences per input, and often they converge slowly, thus requiring large ensembles.

Guo \etal demonstrated that deep neural classifiers are not well calibrated~\cite{guo2017calibration} and proposed temperature scaling for post-hoc calibration of the networks to avoid over-confident or under-confident predictions.
Temperature scaling can be considered a simplification of Platt scaling, where logit values for each category are modeled as Gaussian distributions.
Kull \etal proposed to use Dirichlet distributions to model categorical distributions for each category instead of modeling the logit values using Gaussian distributions.
Mukhoti \etal proposed to use focal loss during training to enhance the calibration of neural networks for classification~\cite{mukhoti2020calibrating}.
Calibration approaches usually use additional in-distribution data for the post-hoc calibration, and the calibrated network can still be miscalibrated for the out-of-distribution data.
Hence, the calibration may not be enough to avoid misleading predictions of neural networks. 

Similarly to \edl, Malinin and Gales proposed Prior Networks~\cite{prior_nets}, which also predict Dirichlet distributions for classification. 
To address the problem of overconfident predictions, Prior Networks utilize an auxiliary dataset that contains out-of-distribution samples. These networks explicitly train neural models to produce highly uncertain outputs for such out-of-distribution samples. However, manually selecting a representative dataset for out-of-distribution samples can be impractical in high-dimensional real-world scenarios due to the vast array of possibilities.

In their subsequent work~\cite{malinin-rkl-2019}, Malinin and Gales recognized significant drawbacks in the original loss function proposed in~\cite{prior_nets}. They opted to adopt \edl's expected cross-entropy loss~\cite{edl}. Unlike probabilistic approaches that rely on sampling methods, \edl and similar methods are deterministic and more practical. They do not require sampling techniques, making them easier to implement and applicable to a wide range of models without architectural changes. Recent studies have also investigated the ability of deterministic neural networks to model predictive uncertainty~\cite{mukhoti2021deterministic, van2020uncertainty, mukhoti2021deep}.

Existing work indicates that \edl has found many applications in many problems.
Ulmer presented a comprehensive overview of single-pass uncertainty quantification methods based on evidential deep learning~\cite{ulmer2021survey}.
Singh \textit{at al.} used evidential deep learning with graph-based clustering for detecting anomalies~\cite{singh2021leveraging}.
Wang \textit{at al.} used uncertainty estimation 
 for stereo matching in computer vision applications~\cite{wang2022uncertainty}.
 Lin \textit{at al.} leveraged evidential neural networks for reliability analysis for finger movement recognition using raw electromyographic signals~\cite{lin2022reliability}. 
 Liu \textit{at al.} empirically showed that \edl's uncertainty score, calculated using total evidence, leads to better performing distance-aware out-of-distribution detection models~\cite{liu2020simple}.

Meinke and Hein proposed a generic model for classifiers with certified low confidence far from their training data~\cite{meinke2019towards}. This model assumes outlier exposure. That is, it requires samples from an out-distribution during training.
Unfortunately, having a comprehensive set of out-of-distribution samples may not be possible while training models in realistic settings.
Unlike methods requiring outlier exposure, the vanilla \edl does not require any out-of-distribution samples during training.
Grcic \textit{at al.} proposed using synthetic negative patches to enhance dense anomaly detection by discriminative training on mixed-content images~\cite{grcic2021dense}.
Sensoy \textit{at al.} also proposed using a generative model to synthesize out-of-distribution training samples to enhance vanilla \edl for out-of-distribution detection~\cite{sensoy2020uncertainty}.
Unlike the methods based on outlier exposure, these approaches exploit generative models to produce out-of-distribution samples.

The most recent uncertainty quantification and robustness approaches rely on large foundational models. 
These models are trained on vast data from various domains; hence, they do not need out-of-distribution samples.
Fort \textit{at al.} demonstrated that large-scale pre-trained transformers could significantly improve the state-of-the-art on a range of out-of-distribution detection tasks across different data modalities ~\cite{fort2021exploring}. They used vision transformers to achieve state-of-the-art results in CIFAR-100 and ImageNet-21k for out-of-distribution detection.
More recently, Tran  \textit{at al.} proposed \textit{Plex}, which is based on pre-trained large models and extension of those using well-known approaches such as efficient ensembling and last layer changes~\cite{tran2022plex}. Plex significantly improves the state-of-the-art across reliability tasks.

\subsection{Risk-aware Classification}

Cost-sensitive learning seeks to build models that assign higher penalties to specific misclassifications~\cite{Elkan01thefoundations,kukar1998cost}. While standard learning methods primarily prioritize classification accuracy, cost-sensitive learning pursues the dual goals of minimizing misclassification costs and achieving high accuracy.

The fusion of cost-sensitive learning with deep neural networks for classification has received limited attention in research. This fusion has given rise to cost-sensitive deep classifiers~\cite{chung2015cost}. Additionally, a direct incorporation of misclassification cost into the cross-entropy loss, using weights derived from the risk matrix, is proposed in~\cite{khan2017cost}.
However, it is important to note that none of these methods consider the uncertainty that comes with classification predictions.

Existing uncertainty quantification methods, including \edl, do not provide any principled way of incorporating misclassification risk into their predictions.
Although uncertainty quantification is essential for autonomous systems using classifiers, risk minimization while making classification decisions under uncertainty is also crucial.
In this paper, we extend \edl by incorporating the risk of misclassification in a principled way.
Furthermore, our approach for risk awareness can also be used by other uncertainty quantification models~\cite{prior_nets, malinin-rkl-2019} based on Dirichlet distributions. 
We borrow ideas from decision theory and integrate the notion of pignistic probabilities \cite{smets_DecisionmakingTBM_05} into neural networks in a novel way using predictive uncertainty.
Our usage of pignistic probabilities in making classification decisions is similar to the use of pignistic probabilities in  Dempster–Shafer theory (DST) of evidence~\cite{dempster2008generalization} and Subjective Logic (SL)~\cite{SLbook}.
However, our \edl-based approaches learn these probabilities using gradient descent as a part of the training of deep neural networks.

\section{Conclusions}
\label{sec:conclusions}
With the increasing adoption of deep learning in a wide range of applications and critical decision-making processes, the challenges related to misclassification are poised to escalate. Consequently, there is a growing demand for the development of \textit{risk-aware classification systems}.
In this paper, we proposed three desiderata to ensure risk-aware classification machinery, \cf \Cref{sec:motivation_full_section}. 
In \Cref{sec:edl_learning}, we revisit the \edl proposal to show that it fulfills the three desiderata. Not only that, it does it simply and elegantly, without \--- for instance \--- requiring external datasets. 
In particular, we demonstrate that \edl loss can be used to fine-tune pre-trained classifiers to improve their uncertainty quantification and turn them into \edl classifiers.
We also propose a principled approach for fusing evidential classifiers trained for different categories and datasets.
We then extend \edl (\cf \Cref{sec:decision}) by incorporating a principled assessment of the misclassification risk, significantly expanding the preliminary results presented at  \cite{DBLP:conf/wacv/SensoySJAR21}, and show how risk-aware \edl classifiers can be trained in environments with bandit feedback.
Our experimental analysis (Section \ref{sec:evaluation}) demonstrates that our approaches significantly outperform the state-of-the-art cost-sensitive training approach for deep learning models.

In future work, we aim to expand our set of desiderata to encompass additional requirements building on our current research interests and expertise, \eg \cite{Guo2024}, notably on three different aspects: adaptability, interpretability of uncertainty, and robustness against adversarial attacks. 
Clearly, encompassing any additional requirements will necessarily require changes in the architectures. We already commented in \Cref{sec:evaluation}, \textbf{riskEDL}, \textbf{EDL(p)}, and \textbf{EDL(pg)} have different strengths and weaknesses: an interesting future direction is to closely link additional requirements with specific neural architecture design patterns.
We plan to study methods for adaptability, allowing classifiers to adjust to new data distributions or environmental changes without significant re-training. Pivotal in this respect could be adopting a causal analysis interpreting the behaviour of the various classifier. 
Providing interpretable uncertainty estimates that are actionable for end-users is another critical area, ensuring that the uncertainty information is both understandable and useful. This involves not only quantifying uncertainty but also making it interpretable and actionable for end-users. Techniques will be developed to present uncertainty in a way that is easily comprehensible, enabling users to make informed decisions based on the classifier’s predictions.
Finally, a major focus will be on enhancing robustness against adversarial attacks: \cite{edl,sensoy2020uncertainty} did demonstrate EDL's robustness against traditional adversarial attacks. However, there might be specialised threat models targeting uncertainty quantification: this would severely impact also any risk-based classification. By incorporating advanced defence mechanisms, we can ensure that the classification systems are resilient to adversarial attacks. This will involve developing techniques to detect and mitigate such attacks, thereby safeguarding the integrity and reliability of the results.

\section*{Acknowledgments}
The work is partially supported by the European Office of Aerospace Research \& Development and the Air Force Office of Scientific Research under award number FA8655-22-1-7017 and by the U.S. Army DEVCOM Army Research Laboratory (DEVCOM ARL) under Cooperative Agreement W911NF2220243. Any opinions, findings, and conclusions or recommendations expressed in this material are those of the author(s) and do not necessarily reflect the views of the United States government.

\section*{Declaration of Generative AI and AI-assisted technologies in the writing process}
During the preparation of this work, the authors used GPT-3.5 \--- OpenAI's large-scale language-generation model \--- to improve readability and language. After using it, the authors reviewed and edited the content as needed and they take full responsibility for the content of the publication.

\bibliography{biblio}
\bibliographystyle{alpha}

\end{document}